\documentclass[10pt,conference,a4paper]{IEEEtran}
\ifCLASSINFOpdf
\else
\fi

\usepackage{times}
\usepackage{epsfig}
\usepackage{graphicx}
\usepackage{amsmath}
\usepackage{amssymb}
\usepackage{hyperref}
\usepackage{graphicx}
\usepackage{amsmath}
\usepackage{xcolor}
\usepackage{amssymb}
\usepackage{subfigure}
\usepackage{multirow}
\usepackage{algorithm2e}
\usepackage{comment}
\usepackage{booktabs}
\usepackage{float}
\usepackage[compress]{cite}

\usepackage[utf8]{inputenc}

\newcommand{\PAR}[1]{\vskip4pt \noindent {\bf #1~}}

\newcommand{\method}{\textbf{CANU}}

\newcommand{\vect}[1]{\mathbf{#1}}

\usepackage{soul}

\usepackage{setspace}

\newcommand{\myto}{\;$\blacktriangleright$}

\begin{document}

%
\title{CANU-ReID: A Conditional Adversarial Network for Unsupervised person Re-IDentification}

\author{\IEEEauthorblockN{Guillaume Delorme$^1$ \quad Yihong Xu$^1$ \quad St\'ephane Lathuili\`ere$^2$ \quad
Radu Horaud$^1$ \quad Xavier Alameda-Pineda$^1$}
\IEEEauthorblockA{$^1$Inria, LJK, Univ. Grenoble Alpes, France \quad $^2$ LTCI, T\'el\'ecom Paris, IP Paris, France\\
Email: firstname.lastname@\{$^1$inria.fr \quad $^2$telecom-paris.fr\}}
}


%


\pagestyle{plain}
\twocolumn[{%
\renewcommand\twocolumn[1][]{#1}%
\maketitle
\thispagestyle{plain}
\pagestyle{plain}
\begin{center}
    \centering
    \includegraphics[width=0.8\textwidth]{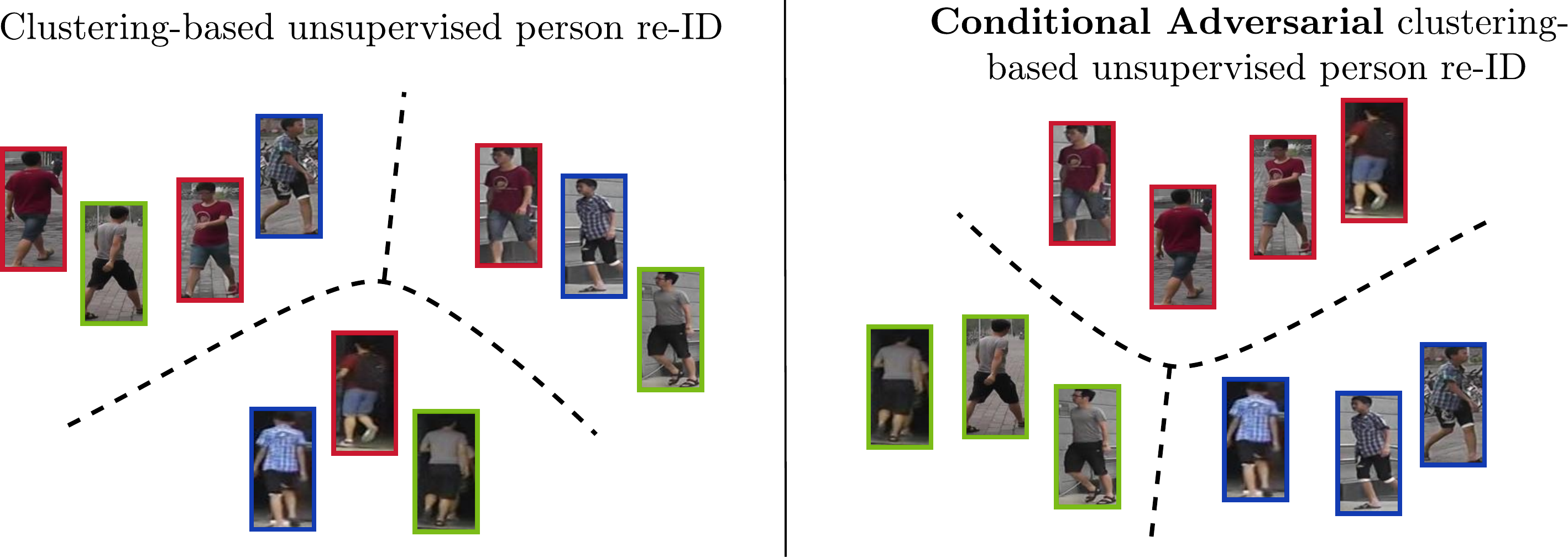}\\
\end{center}
{\footnotesize

Fig. 1: Clustering-based (left) vs.\ conditional adversarial clustering-based (right) unsupervised person re-ID. Our intuition is that features should be camera-independent, and thus the clustering result should group visual features by ID rather than by camera. Our method conditions a camera-based adversarial discriminator with the visual features corresponding to the cluster's centroid.

}\stepcounter{figure}
}]

\begin{abstract}Unsupervised person re-ID is the task of identifying people on a target data set for which the ID labels are unavailable during training. In this paper, we propose to unify two trends in unsupervised person re-ID: clustering \& fine-tuning and adversarial learning. On one side, clustering groups training images into pseudo-ID labels, and uses them to fine-tune the feature extractor. On the other side, adversarial learning is used, inspired by domain adaptation, to match distributions from different domains. Since target data is distributed across different camera viewpoints, we propose to model each camera as an independent domain, and aim to learn domain-independent features. Straightforward adversarial learning yields negative transfer, we thus introduce a conditioning vector to mitigate this undesirable effect. In our framework, the centroid of the cluster to which the visual sample belongs is used as conditioning vector of our conditional adversarial network, where the vector is permutation invariant (clusters ordering does not matter) and its size is independent of the number of clusters. To our knowledge, we are the first to propose the use of conditional adversarial networks for unsupervised person re-ID. We evaluate the proposed architecture on top of two state-of-the-art clustering-based unsupervised person re-identification (re-ID) methods on four different experimental settings with three different data sets and set the new state-of-the-art performance on all four of them. Our code and model will be made publicly available at \url{https://team.inria.fr/perception/canu-reid/}. \end{abstract}


%
\IEEEpeerreviewmaketitle

\section{Introduction}

Person re-identification (re-ID) is a well-studied retrieval task~\cite{10.1109/ICPR.2014.16,385d6453d2f04126994b710df0df5c96,8545620} that consists in associating images of the same person across cameras, places and time. Given a query image of a person, we aim to recover his/her identity (ID) from a set of identity-labeled gallery images. The person re-ID task is particularly challenging for two reasons. First, query and gallery images contain only IDs which have never been seen during training.
Second, gallery and query images are captured under a variety of background, illumination, viewpoints and occlusions.

Most re-ID models assume the availability of heavily labeled datasets and focus on improving their performance on the very same data sets, see for instance~\cite{DUKEMTMC,Market}. The limited generalization capabilities of such methods were pointed out in previous literature~\cite{deng2018image,10.1145/3243316}. In the recent past, researchers attempted to overcome this limitation by investigating a new person re-ID task, where there is a \textit{source} dataset annotated with person IDs and another unlabeled \textit{target} dataset. This is called \textit{unsupervised} person re-ID. Roughly speaking, the current trend is to use a pre-trained base architecture to extract visual features, cluster them, and use the cluster assignments as \textit{pseudo-labels} to re-train the base architecture using standard supervised re-ID loss functions~\cite{fu2019self,ge2020mutual}.

In parallel, since generative adversarial networks were proposed, adversarial learning has gained popularity in the domain adaptation field~\cite{Tzeng, ganin2016domain, domain_sep_net}. The underlying intuition is that learning a feature generator robust to the domain shift between \textit{source} and \textit{target} would improve the target performance. The adversarial learning paradigm has been successfully used for person re-ID in both the supervised~\cite{CamSty,DUKE_REID}, and the unsupervised~\cite{10.1145/3243316,lin2019bottom} learning paradigms.

In this paper, we propose to unify these two trends in unsupervised person re-ID: hence using conditional adversarial networks for unsupervised person re-ID. Our intuition is that good person re-ID visual features should be independent of the camera/viewpoint, see Fig.~1. Naturally, one would expect that an adversarial game between a generator (feature extractor) and a discriminator (camera classifier) should suffice. However, because the ID presence is not uniform in all cameras, such simple strategy implies some negative transfer and limits -- often decreases -- the representational power of the visual feature extractor. To overcome this issue, we propose to use conditional adversarial networks, thus providing an additional identity representation to the camera discriminator.
Since in the target dataset, the ID labels are unavailable, we exploit the pseudo-labels. More precisely, we provide, as conditioning vector, the centroid of the cluster to which the image belongs. The contributions of this paper are the following:
\begin{itemize}
    \item We investigate the impact of a camera-adversarial strategy in the unsupervised person re-ID task.
    \item We realize the negative transfer effect, and propose to use conditional adversarial networks.
    \item The proposed method can be easily plugged into any unsupervised clustering-based person re-ID methods. We experimentally combine \method~with two clustering-based unsupervised person re-ID methods, and propose to use their cluster centroids as conditioning labels.
    \item Finally, we perform an extensive experimental validation on four different unsupervised re-ID experimental settings and outperform current state-of-the-art methods by a large margin on all settings.
\end{itemize}

The rest of the paper is organized as follows. Section~\ref{sec:sota} describes the state-of-the-art. Section~\ref{sec:method-back} discusses the basics of clustering-based unsupervised person re-ID and sets the notations. The proposed conditional adversarial strategy is presented in Section~\ref{sec:canU}. The extensive experimental validation is discussed in Section~\ref{sec:experimental-validation} before drawing the conclusions in Section~\ref{sec:conclusions}.
\section{Related work}
\label{sec:sota}





\PAR{Unsupervised person re-identification (re-ID)} has drawn growing attention in the last few years, taking advantage of the recent achievements of supervised person re-ID models, without requiring an expansive and tedious labeling process of the target data set. A very important line of research starts from a pre-trained model on the source data set and is based on \textit{clustering} and \textit{fine-tuning}~\cite{10.1145/3243316,lin2019bottom,fu2019self,ge2020mutual,zhang2019self}. It alternates between a clustering step generating noisy pseudo-labels, and a fine-tuning step adapting the network to the target data set distribution, leading to a progressive label refinement. Thus, these methods do not use the source data set during adaptation. A lot of effort has been invested in improving the quality of the pseudo-labels. Sampling from reliable clusters during adaptation~\cite{10.1145/3243316}, gradually reducing the number of clusters and merging by exploiting intrinsic inter-ID diversity and intra-ID similarity~\cite{lin2019bottom}, or performing multiple clustering on visual sub-domains and enforcing consistency~\cite{fu2019self} have been investigated. More recently, \cite{ge2020mutual} investigated the interaction of two different models to assess and incorporate pseudo-label reliability within a teacher-student framework. 

A different approach is directly inspired by Unsupervised Domain Adaptation (UDA)~\cite{deng2018image,zhong2018generalizing,chang2019disjoint,qi2019novel,song2020unsupervised,zhong2019invariance}: using both the source and target data sets during adaptation. These methods aim to match the distributions on the two data sets while keeping its discriminative ability leveraging source ground truth ID labels. A first strategy learns to map source's detections to target's style detections, and train a re-ID model in a supervised setting using those only those transferred detections~\cite{deng2018image}, or in combination with the original target detections~\cite{zhong2018generalizing}. More standard UDA strategies use adversarial learning to match the source and target distributions~\cite{ganin2016domain,qi2019novel}.



\PAR{Negative transfer} has been investigated in unsupervised domain adaptation~\cite{torrey.handbook09}, especially for Partial Domain Adaptation (PDA)~\cite{ETN_2019_CVPR,SAN,IWAN}, where target labels are only a subset of the source's. Negative transfer is defined as the inability of an adaptation method to find underlying common representation between data sets and is generally caused by the gap between the distributions of the two data sets being too wide~\cite{wang2019characterizing} for the algorithm to transfer knowledge. Weighting mechanisms are generally employed to remove the impact of source's outliers class on the adaptation process, either for the matching part ~\cite{IWAN,SAN,STN}, the classification part~\cite{wang2019characterizing}, or both~\cite{ETN_2019_CVPR}. Interestingly, \cite{wang2019characterizing} uses a domain discriminator conditioned by source label to perform conditional distribution matching. Investigating negative transfer is not limited to UDA settings. For example, a similar method has been proposed for domain generalization~\cite{Li_2018_ECCV}, implementing a conditional discriminator to match conditioned domain distributions. By doing so, the impact of the difference between prior label distributions on the discriminative ability of the model is alleviated.

\begin{figure*}
\vspace*{+2mm}
\centering
\includegraphics[width=\textwidth]{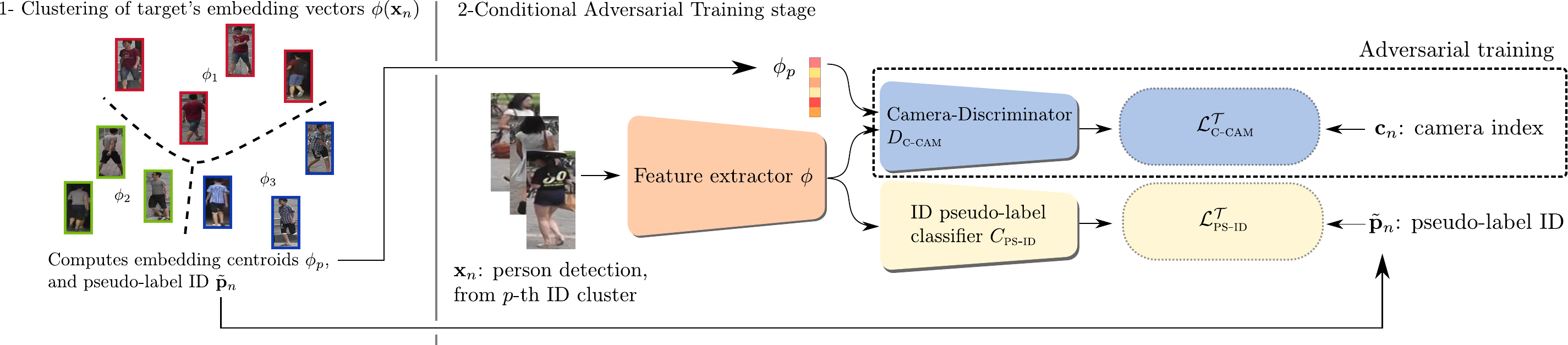}
\vspace*{-5mm}
\caption{Pipeline of our method: alternatively (1) clustering target's training data set using $\phi$ representation, producing noisy pseudo-label ID $\tilde{p}_n$ alongside centroids $\phi_p$, and (2) conditional adversarial training, using a Camera-Discriminator $D_{CAM}$ conditioned by $\phi_p$ to enforce camera invariance on a per identity basis to avoid negative transfer. Pseudo-label ID are used to train an ID classifier $C_{PS-ID}$ alongside the discriminator. \label{fig:arch}}
\vspace*{-5mm}
\end{figure*}

Within the task of unsupervised person re-ID, different cameras could be considered as different domains, and standard matching strategies could be used. However, they would inevitably induce negative transfer as described before for generic domain adaptation. Direct application of PDA methods into the person re-ID tasks is neither simple nor expected to be successful. The main reason is that, while PDA methods handle a few dozens of classes, standard re-ID data sets contain a few thousands of IDs. This change of scale requires a different strategy, and we propose to use conditional adversarial networks, with a conditioning label that describes the average sample in the cluster, rather than representing the cluster index. In conclusion, different from clustering and fine-tuning unsupervised person re-ID methods, we propose to exploit (conditional) adversarial networks to learn visual features that are camera independent and thus more robust to appear changes. Different from previous domain adaptation methods, we propose to match domains (cameras) with a conditioning label that evolves during training, since it is the centroid of the cluster to which the visual sample is assigned, allowing us having a representation that is independent of the number of clusters and the cluster index.
\section{Clustering based Unsupervised Person Re-ID}
\label{sec:method-back}
We propose to combine conditional adversarial networks with clustering-based unsupervised person Re-ID. To detail our contributions, we first set up the basics and notations of existing methods for unsupervised person re-ID.

Let $\cal S$ denote a source ID-annotated person re-ID dataset, containing $N^{\cal S}$ images corresponding to $M^{\cal S}$ different person identities captured by $K^{\cal S}$ cameras. We write ${\cal S}=\{(\vect{x}^{\cal S}_n,\vect{p}_n^{\cal S},\vect{c}_n^{\cal S})\}_{n=1}^{N^{\cal S}}$, where each three-tuple consists of a detection image, $\vect{x}^{\cal S}_n$, a person ID one-hot vector, $\vect{p}_n^{\cal S}\in\{0,1\}^{M^{\cal S}}$ and a camera index one-hot vector, $\vect{c}_n^{\cal S}\in\{0,1\}^{K^{\cal S}}$. Similarly, we define ${\cal T} =\{(\vect{x}^{\cal T}_n,\vect{c}_n^{\cal T})\}_{n=1}^{N^{\cal T}}$ a target person re-ID dataset, with $K^{\cal T}$ cameras and $N^{\cal T}$ element, without ID labels.

\PAR{Source pre-training} Let  $\phi$  be a convolutional neural network backbone (\textit{e.g.} ResNet-50~\cite{ResNet}) 
served as a trainable \textit{feature extractor}. The goal of person re-ID is to be able to discriminate person identities, and therefore an identity classifier $C_{\textsc{id}}$ is required. The output of $C_{\textsc{id}}$ is a $M^{\cal S}$-dimensional stochastic vector, encoding the probability of the input to belong to each of the identities. The cross-entropy and triplet losses are usually employed:
\begin{equation}
{\cal L}_\textsc{ce}^{\cal S}(\phi,C_\textsc{id}) = - \mathbb{E}_{(\vect{x}^{\cal S},\vect{p}^{\cal S})\sim{\cal S}} \left\{ \log \left< C_\textsc{id}(\phi(\vect{x}^{\cal S})), \vect{p}^{\cal S}\right> \right\},\!\!\!\label{eq:lossID}\vspace{-2mm}
\end{equation}
\begin{align}
{\cal L}_\textsc{tri}^{\cal S}(\phi) = &\mathbb{E}_{(\vect{x}^{\cal{S}},\vect{x}_p^{\cal{S}},\vect{x}_n^{\cal{S}})\sim{\cal P}_{\cal{S}}} \{\max(0,\| \phi(\vect{x}^{\cal{S}}) - \phi(\vect{x}_p^{\cal{S}}) \| \nonumber \\
& +m - \| \phi(\vect{x}^{\cal{S}}) - \phi(\vect{x}_n^{\cal{S}}) \|)\}, 
\end{align}
where $\mathbb{E}$ denotes the expectation, $\left<\cdot,\cdot\right>$ the scalar product, $\| . \|$ the $L^2$-norm distance, $\vect{x}_p^{\cal{S}}$ and $\vect{x}_n^{\cal{S}}$ are the hardest positive and negative example for $\vect{x}^{\cal{S}}$ in ${\cal P}_{\cal{S}}$ the set of all triplets in $\cal{S}$, and $m = 0.5$. We similarly denote ${\cal L}_\textsc{ce}^{\cal T}$ and ${\cal L}_\textsc{tri}^{\cal S}$ the cross-entropy and triplet losses evaluated on the target dataset. However, in unsupervised reID settings, target ID labels are unavailable, and therefore we will need to use alternative \textit{pseudo-ID labels}.
The re-ID feature extractor $\phi$ is typically trained using:
\begin{align}
 {\cal L}_\textsc{id}^{\cal S}(\phi,C_\textsc{id}) = {\cal L}_\textsc{ce}^{\cal S}(\phi,C_\textsc{id}) + \lambda {\cal L}_\textsc{tri}^{\cal S}(\phi),\label{eq:full-loss}
\end{align}
for a fixed balancing value $\lambda$, achieving competitive performance on the source test set \cite{DBLPHermansBL17}. However, they notoriously lack generalization power and perform badly on datasets unseen during training~\cite{deng2018image}, thus requiring adaptation.

\PAR{Target fine-tuning} As discussed above, target ID labels are unavailable. To overcome this while leveraging the discriminative power of widely-used losses described in Eq.~\ref{eq:full-loss}, methods like~\cite{fu2019self,ge2020mutual} use pseudo-labels. 
The hypothesis of these methods is that the features learned during the pre-training stage are exploitable for the inference of target's ID labels to a certain extent. Starting from the pre-trained model, these methods alternate between (i) pseudo ID label generation $\{{\tilde{\vect{p}}^{\cal T}_n}\}_{n=1}^{N^{\cal T}}$ using a standard clustering algorithm (k-means or DBSCAN~\cite{DBSCAN}) on the target training set $\{\phi(\vect{x}^{\cal T}_n)\}_{n=1}^{N^{\cal T}}$ and (ii) the update of $\phi$ using losses similar to Eq.~\ref{eq:full-loss} supervised by $\{{\tilde{\vect{p}}^{\cal T}_n}\}_{n=1}^{N^{\cal T}}$. 
Since our approach is agnostic to the ID loss used at this step, we choose to denote it by ${\cal L}_{\textsc{ps-id}}(\phi,C_\textsc{ps-id})$, $C_\textsc{ps-id}$ being an optional classifier layer for the pseudo-labels, and develop it further in the experimental section.

\section{CANU-ReID: A Conditional Adversarial Network for Unsupervised Person re-ID}
\label{sec:canU}
In this section we discuss the main limitation of clustering-based unsupervised re-ID methods: we hypothesize that viewpoint variability can make things difficult for clustering methods and propose two alternatives. First, an adversarial network architecture targeting re-ID features that are camera-independent. This strategy could, however, induce some negative transfer when the correlation between cameras and IDs is strong. Second, a conditional adversarial network architecture specifically designed to overcome this negative transfer.

\PAR{Camera adversarial-guided clustering} We hypothesize that camera (viewpoint) variability is one of the major limiting factors for clustering-based unsupervised re-ID methods. In plain, if the embedding space variance explained by camera changes is high, the clustering method could be clustering images from the same camera, rather than images from the same ID. Therefore, $\phi$ will produce features that can very well discriminate the camera at the expense of the ID. 
To alleviate this problem, we propose to directly enforce camera invariance in $\phi$'s representation by using an adversarial strategy, where the discriminator is trained to recognize the camera used to capture the image. Consequently, the generator, in our case $\phi$, is trained to remove any trace from the camera index (denoted by $\mathbf{c}$). Intuitively, this should reduce the viewpoint variance in the embedding space, improve pseudo-labels quality and increase the generalization ability of $\phi$ to unseen IDs.\\

To do so, we require a camera discriminator $D_\textsc{cam}$ (see Fig.~\ref{fig:arch} for a complete overview of the architecture). The generator $\phi$ and the discriminator $D_\textsc{cam}$ will be trained through a min-max formulation:
\begin{equation}
 \min_{\phi,C_\textsc{ps-id}} \max_{D_\textsc{cam}} {\cal L}_\textsc{ps-id}^{\cal T}(\phi,C_\textsc{ps-id}) - \mu{\cal L}_\textsc{cam}^{{\cal T}}(\phi,D_\textsc{cam}),\label{eq:full_adv}
\end{equation}
where $\mu>0$ is a balance hyper-parameter that can be interpreted as a regularization parameter~\cite{ganin2016domain}, and ${\cal L}_\textsc{cam}^{{\cal T}}$ is defined via the cross-entropy loss:
\begin{equation}
 {\cal L}_\textsc{cam}^{{\cal T}}(\phi,D_\textsc{cam}) = - \mathbb{E}_{(\vect{x}^{\cal{T}},\vect{c}^{\cal{T}})\sim{{\cal T}}} \left\{ \log \left< D_\textsc{cam}(\phi(\vect{x}^{\cal{T}})), \vect{c}^{\cal{T}}\right> \right\}
\end{equation}

On one side, the feature extractor $\phi$ must minimize the person re-ID loss ${\cal L}_\textsc{ps-id}$ at the same time as making the problem more challenging for the camera discriminator. On the other side, the camera discriminator tries to learn to recognize the camera corresponding to the input image.

\PAR{Adversarial negative transfer} It has been shown~\cite{Li_2018_ECCV} that minimizing~(\ref{eq:full_adv}) is equivalent to the following problem:
\begin{align}
\label{eq:JSD}
\displaystyle\min_{\phi,C_\textsc{ps-id}} & \; {\cal L}_\textsc{ps-id}^{\cal T}(\phi,C_\textsc{ps-id}) \\
\text{s.t.} & \; \textrm{JSD}_{{\cal T}}(p(\phi(\vect{x})|\vect{c}=1),\ldots,p(\phi(\vect{x})|\vect{c}=K)) = 0, \nonumber
\end{align}
where $\textrm{JSD}_{{\cal T}}$ stands for the multi-distribution Jensen-Shanon divergence~\cite{JSD} on the target set ${\cal T}$, and we dropped the superscript ${\cal T}$ in the variables to ease the reading.

Since the distribution of ID labels may strongly depend on the camera, the plain adversarial strategy in~(\ref{eq:JSD}) can introduce negative transfer~\cite{wang2019characterizing}. Formally, since we have:
$$p(\vect{p}|\vect{c}=i) \neq p(\vect{p}|\vect{c}=j), i\neq j$$
then solving~(\ref{eq:JSD}) is not equivalent (see~\cite{Li_2018_ECCV}) to:
\begin{align}
\label{eq:JSD_cond}
\displaystyle\min_{\phi,C_\textsc{ps-id}} & {\cal L}_\textsc{ps-id}^{\cal T}(\phi,C_\textsc{ps-id}) \\
\text{s.t.} & \; \textrm{JSD}_{{\cal T}}(p(\phi(\vect{x})|\vect{p}, \vect{c}=1), \ldots , p(\phi(\vect{x})| \vect{p}, \vect{c}=K)) = 0, \nonumber
\end{align}
which is the problem we would implicitly want to solve. Intuitively, \textit{negative transfer} means that the camera discriminator learns $p(\vect{c}|\vect{p})$ instead of $p(\vect{c}|\vect{x},\vect{p})$, exploiting ID to infer camera information and decreasing the representation power of $\phi$ due to the adversarial loss.\\

\PAR{Conditional adversarial networks} We propose to directly solve the optimization problem in Eq.~\ref{eq:JSD_cond} to alleviate the negative transfer. Similar to the original conditional GAN formulation~\cite{COND_GAN}, we condition the adversarial discriminator with the input ID $\vect{p}$. Given that ID labels are unavailable on the target set, we replace them by the pseudo-labels obtained during the clustering phase.

However, since we are handling a large number of IDs (700 to 1500 in standard re-ID datasets), using a one-hot representation turned out to be very ineffective. Indeed, such representation is not permutation-invariant, meaning that if the clusters are re-ordered, the associated conditional vector changes, which does not make sense. We, therefore, need a permutation-invariant conditioning label.


To do so, we propose to use the cluster centroids $\vect{\phi}_{\vect{p}}$ which are provided by the clustering algorithms at no extra cost. This conditioning vectors are permutation invariant. Importantly, we do not back-propagate the adversarial loss through the ID-branch, to avoid using an ID-dependant gradient from the adversarial loss. This boils down to defining  ${\cal L}_\textsc{c-cam}$ as:
\begin{equation}
 {\cal L}_\textsc{c-cam}^{{\cal T}}(\phi,D_\textsc{c-cam}) =  - \mathbb{E}_{(\vect{x}, \vect{p}, \vect{c})\sim{{\cal T}}} \left\{ \log \left< D_\textsc{c-cam}(\phi(\vect{x}) ,\vect{\phi}_{\vect{p}}), \vect{c}\right> \right\} \label{eq:cond-cam}
\end{equation}
and then solving:
\begin{equation}
 \min_{\phi,C_\textsc{ps-id}} \max_{D_\textsc{c-cam}} {\cal L}_\textsc{ps-id}^{\cal T}(\phi,C_\textsc{ps-id}) - \mu{\cal L}_\textsc{c-cam}^{{\cal T}}(\phi,D_\textsc{c-cam}).\label{eq:full_c-adv}
\end{equation}
\section{Experimental Validation}
\label{sec:experimental-validation}
In this section, we provide implementation details and an in-depth evaluation of the proposed methodology, setting the new state-of-the-art in four different unsupervised person re-ID experimental settings. We also provide an ablation study and insights on why conditional adversarial networks outperform existing approaches.
\subsection{Evaluation Protocol}
\label{subsec:experimentalsetup}
We first describe here the baselines, on which our proposed \method~is built and tested. The used datasets and the evaluation metrics are then introduced.
\PAR{Baselines} 
The proposed~\method~can be easily plugged into any clustering-based unsupervised person re-ID methods. Here, we experimentally test it on two state-of-the-art clustering-based unsupervised person re-ID methods, as baselines. 

First, self-similarity grouping~\cite{fu2019self} (\textbf{SSG}) performs independent clustering on the upper-, lower- and full-body features, denoted as $\phi^{\textsc{u}}$, $\phi^{\textsc{l}}$ and $\phi^{\textsc{f}}$. They are extracted from three global average pooling layers of the convolutional feature map of ResNet-50~\cite{ResNet}. The underlying hypothesis is that noisy global pseudo-label generation can be improved by using multiple, but related clustering results, and enforcing consistency between them. The triplet loss is used to train the overall architecture.

To implement \method-SSG, we define three different camera discriminators, one for each embedding, $D_{\textsc{c-cam}}^{\textsc{u}}$, $D_{\textsc{c-cam}}^{\textsc{l}}$ and $D_{\textsc{c-cam}}^{\textsc{f}}$ respectively, each fed with samples from the related representation and conditioned by the global embedding $\phi^{\textsc{f}}$. In the particular case of \method-SSG, the generic optimisation problem in Eq.~\ref{eq:full_c-adv} instantiates as:\vspace{-4mm}

\begin{align}
\label{eq:ssg_ad}
 \displaystyle\min_{\phi} \max_{D_\textsc{c-cam}^{\textsc{u},\textsc{l},\textsc{f}}} & {\cal L}_\textsc{ssg}^{\cal T}(\phi) 
 - \mu{\cal L}_\textsc{c-cam}^{{\cal T}}(\phi^{\textsc{u}},D_\textsc{c-cam}^{\textsc{u}}) \\ 
 &- \mu{\cal L}_\textsc{c-cam}^{{\cal T}}(\phi^{\textsc{l}},D_\textsc{c-cam}^{\textsc{l}}) 
 - \mu{\cal L}_\textsc{c-cam}^{{\cal T}}(\phi^{\textsc{f}},D_\textsc{c-cam}^{\textsc{f}}). \nonumber
\end{align}

Second, Mutual Mean-Teaching~\cite{ge2020mutual} (\textbf{MMT}) reduces pseudo-label noise by using a combination of hard and soft assignment: using hard labeling reduces the amount of information given to the model, and using soft labeling allows the cluster's confidence to be taken into account. MMT defines two different models $(\phi^1,C_\textsc{ps-id}^1)$ and $(\phi^2,C_\textsc{ps-id}^2)$, both implemented with a IBN-ResNet-50~\cite{pan2018two} backbone, initialized with two different pre-trainings on the source dataset. They are then jointly trained using pseudo labels as hard assignments, and inspired by teacher-student methods, using their own pseudo ID predictions as soft pseudo-labels to supervise each other. Soft versions of cross-entropy and triplet loss are used.

To implement \method-MMT, similar to \method-SSG, we define two camera discriminators $D_{\textsc{c-cam}}^1$ and $D_{\textsc{c-cam}}^2$, each dedicated to one embedding, and train it using the following instantiation of the generic optimisation problem in Eq.~\ref{eq:full_c-adv}:\vspace{-4mm}

\begin{align}
\label{eq:mmt_ad}
 \min_{\phi^{1,2},C_\textsc{ps-id}^{1,2}} \max_{D_\textsc{c-cam}^{1,2}} &{\cal L}_\textsc{mmt}^{\cal T}(\phi^1,C_\textsc{ps-id}^1) + {\cal L}_\textsc{mmt}^{\cal T}(\phi^2,C_\textsc{ps-id}^2) \\ 
 &- \mu{\cal L}_\textsc{c-cam}^{{\cal T}}(\phi^1,D_\textsc{c-cam}^1) - \mu{\cal L}_\textsc{c-cam}^{{\cal T}}(\phi^2,D_\textsc{c-cam}^2). \nonumber
\end{align}

While the clustering strategy used in SSG is DBSCAN~\cite{DBSCAN}, the one used in MMT is standard k-means. For a fair comparison, we implemented \method~with DBSCAN, which has the advantage of automatically selecting the number of clusters. We also evaluate the performance of MMT using the DBSCAN clustering strategy without \method, to evaluate the impact of our method on a fair basis.

\PAR{Datasets}
The proposed adversarial strategies are evaluated using three datasets: Market-1501 (Mkt)~\cite{Market}, DukeMTMC-reID (Duke)~\cite{DUKEMTMC} and MSMT17 (MSMT)~\cite{Wei_2018_CVPR}. In all three cases, the dataset is divided into three parts: training, gallery, and query. The query and the gallery are never available during training and only used for testing. 

Mkt is composed of $M=1,501$ (half for training and half for testing) different identities, observed through $K=6$ different cameras (viewpoints). The deformable parts model~\cite{DPM} is used for person detection. As a consequence, there are $N=12,936$ training images and $19,732$ gallery images. In addition, there are $3,368$ hand-drawn bounding box queries.

Duke is composed of $M=1,404$ (half for training and half for testing) identities captured from $K=8$ cameras. In addition, $408$ other ID, called ``distractors'', are added to the gallery. Detections are manually selected, leading to $N=16,522$ images for train, $17,661$ for the gallery and $2,228$ queries.

MSMT is the largest and most competitive dataset available, with $M=4,101$ identities ($1,041$ for training, and $3,060$ for test), $K=15$ cameras, with $N=32,621$ images for training, $82,161$ for the Gallery and $11,659$ queries.

The unsupervised person re-ID experimental setting using dataset A as source and dataset B as the target is denoted by A\myto\ B. We compare the proposed methodology in four different settings: Mkt\myto\ Duke, Duke\myto\ Mkt, Mkt\myto\ MSMT and Duke\myto\ MSMT.

\PAR{Evaluation metrics}
In order to provide an objective evaluation of the performance, we employ two standard metrics in person re-ID~\cite{Market}: Rank-1 (R1) and mean average-precision (mAP). Precisely, for each query image, we extract visual features employing $\phi$, and we compare them to the features extracted from the gallery using the cosine distance. Importantly, the gallery images captured with the same camera as the query image are not considered. For R1, a query is well identified if the closest gallery feature vector corresponds to the same identity. In the case of mAP, the whole list of gallery images is considered, and precision at different ranking positions is averaged. See~\cite{Market} for details. For both metrics, the mean over the query set is reported.


\PAR{Implementation details}
For both MMT and SSG, we use the models pre-trained on the source datasets (e.g. For Mkt\myto Duke, we use the model pre-trained on the Market dataset and provided by~\cite{fu2019self} and \cite{ge2020mutual}). DBSCAN is used at the beginning of each training epoch, the parameters for DBSCAN are the same described as in \cite{fu2019self}. The weight for (conditional) adversarial losses $\mu$ is set to $0.1$ for MMT and to $0.05$ for SSG, chosen according to a grid search with values between $[0.01, 1.8]$ (see below). The used conditional discriminator has two input branches, one as the (conditional) ID branch and the other is the camera branch, both consist of four fully-connected layers, of size $[2048,1024]$, $[2048,1024]$, $[1024, 1024]$, $[1024, \text{number of cameras}]$, respectively. Batch normalization~\cite{ioffe2015batch} and ReLU activation are used. For MMT, during the unsupervised learning, we train the IBN-ResNet-50~\cite{pan2018two} feature extractor with Adam~\cite{adam} optimizer using a learning rate of $0.00035$. As default in~\cite{ge2020mutual}, the network is trained for 40 epochs but with fewer iterations per epoch (400 v.s.\, 800 iterations) while keeping a similar or better performance. For SSG, we train the ResNet-50~\cite{ResNet} with SGD optimizer using a learning rate of $6e$-$5$. At each epoch, unlike MMT, we iterate through the whole training set instead of training with a fix number of iterations. 

After training, the discriminator is discarded and only the feature extractor is kept for evaluations. For SSG, first, it combines the features extracted from the original image and the horizontally flipped image with a simple sum. Second, the summed features are normalized by their $L_2$ norm. Finally, The full-, upper- and, lower-body normalized features are concatenated to form the final features. For MMT, the features extracted from the feature extractor are directly used for evaluations.

In the following, we first compare the proposed methodology with the state-of-the-art~(see Sec.~\ref{subsec:sota}). Secondly, we discuss the benefit of using conditional camera-adversarial training in the ablation study (see Sec.~\ref{subsec:ablation}), and include several insights on the performance of \method.
\begin{table}[t]
\caption{Comparison of the proposed \textbf{\method} methodology on the Mkt\myto\ Duke and Duke\myto\ Mkt unsupervised person re-ID settings. \textbf{\method-MMT} establishes a new state-of-the-art in both settings, and \textbf{\method-SGG} outperforms \textbf{SSG}.}\vspace{-3mm}
\label{tab:sota-Duke-Mkt}
        \centering
        \begin{tabular}{lcccc}
        \toprule
        \multirow{2}{*}{Method} & \multicolumn{2}{c}{Mkt\myto\ Duke} & \multicolumn{2}{c}{Duke\myto\ Mkt} \\
        \cmidrule{2-5} 
         & R1 & mAP & R1 & mAP \\
        \midrule
        PUL~\cite{10.1145/3243316}              & 30.0  & 16.4  & 45.5  & 20.5 \\
        TJ-AIDL~\cite{wang2018transferable}     & 44.3  & 23.0  & 58.2  & 26.5 \\
        SPGAN~\cite{deng2018image}              & 41.1  & 22.3  & 51.5  & 22.8 \\
        HHL\cite{zhong2018generalizing}         & 46.9  & 27.2  & 62.2  & 31.4 \\
        CFSM~\cite{chang2019disjoint}           & 49.8  & 27.3  & 61.2  & 28.3 \\
        BUC~\cite{lin2019bottom}                & 47.4  & 27.5  & 66.2  & 38.3 \\
        ARN~\cite{li2018adaptation}             & 60.2  & 33.4  & 70.3  & 39.4 \\
        UDAP~\cite{song2020unsupervised}        & 68.4  & 49.0  & 75.8  & 53.7 \\
        ENC~\cite{zhong2019invariance}          & 63.3  & 40.4  & 75.1  & 43.0 \\
        UCDA-CCE~\cite{qi2019novel}             & 47.7  & 31.0  & 60.4  & 30.9 \\
        PDA-Net~\cite{li2019cross}              & 63.2  & 45.1  & 75.2  & 47.6 \\
        PCB-PAST~\cite{zhang2019self}           & 72.4  & 54.3  & 78.4  & 54.6 \\
        Co-teaching~\cite{han2018co}            & 77.6  & 61.7  & 87.8  & 71.7 \\
        \midrule
        SSG~\cite{fu2019self}                   & 73.0  & 53.4  & 80.0  & 58.3 \\
        \method-SSG (ours)                      & 76.1  & 57.0  & 83.3  & 61.9 \\ 
        \midrule
        MMT~\cite{ge2020mutual}                 & 81.8  & 68.7  & 91.1  & 74.5 \\
        MMT (DBSCAN)                            & 80.2  & 67.2  & 91.7  & 79.3 \\
        \method-MMT (ours) & \textbf{83.3} & \textbf{70.3} & \textbf{94.2} & \textbf{83.0}\\
        \bottomrule
    \end{tabular}
\end{table}

\subsection{Comparison with the State-of-the-Art}
\label{subsec:sota}
We compare \method-SSG and \method-MMT to the state-of-the-art methods and we demonstrate in Tables~\ref{tab:sota-Duke-Mkt} and~\ref{tab:sota-MSMT} that \method-MMT sets a new state-of-the-art result compared to the existing unsupervised person re-ID methods by a large margin. In addition, \method-SSG outperforms SSG in all settings. Since the MSMT dataset is more recent, fewer comparisons are available in the experiments involving this dataset, hence the two different tables.

More precisely, the proposed \method~significantly improves the performance of the baselines, SSG~\cite{fu2019self} and MMT~\cite{ge2020mutual}. In Mkt\myto\!\! Duke and Duke\myto\!\! Mkt (Table~\ref{tab:sota-Duke-Mkt}), \method-SSG improves SSG by $\uparrow$3.1\%/$\uparrow$3.6\% (R1/mAP, same in the following.) and $\uparrow$3.3\%/$\uparrow$3.6\% respectively, and \method-MMT significantly outperforms MMT by $\uparrow$1.5\%/$\uparrow$1.6\% and $\uparrow$3.1\%/$\uparrow$8.5\% respectively. Moreover, for the more challenging setting (Table~\ref{tab:sota-MSMT}), the improvement brought by \method~is even more evident. For SSG, for example, we increase the R1/mAP by $\uparrow$13.9\%/$\uparrow$5.9\% in Mkt\myto\!\! MSMT, and by $\uparrow$11.1\%/$\uparrow$4.6\% in Duke\myto\!\! MSMT. For MMT, \method-MMT outperforms MMT by $\uparrow$7.3\%/$\uparrow$8.0\% in Mkt\myto\!\! MSMT, and by $\uparrow$8.7\%/$\uparrow$9.0\% in Duke\myto\!\! MSMT. Finally, the consistent improvement in the four settings of \method-MMT over MMT (DBSCAN) and the inconsistent improvement of MMT (DBSCAN) over standard MMT proves that the increase of the performance is due to the proposed methodology.
To summarize, we greatly improve the baselines using the proposed \method. More importantly, to our best knowledge, we outperform the existing methods by a large margin and establish a new state-of-the-art result.

\begin{table}[t]
\centering
\caption{Comparison of the proposed \textbf{\method} methodology on the Mkt\myto\ MSMT and Duke\myto\ MSMT unsupervised person re-ID settings. \textbf{\method-MMT} establishes a new state-of-the-art in both settings, and \textbf{\method-SGG} outperforms \textbf{SSG}.}\vspace{-3mm}
\label{tab:sota-MSMT}
        \begin{tabular}{lcccc}
        \toprule
        \multirow{2}{*}{Method}  & \multicolumn{2}{c}{Mkt\myto\ MSMT} & \multicolumn{2}{c}{Duke\myto\ MSMT} \\
        \cmidrule{2-5} 
        & R1 & mAP & R1 & mAP \\
        \midrule
        PTGAN~\cite{wei2018person}      & 10.2 &  2.9 & 11.8 &  3.3  \\
        ENC~\cite{zhong2019invariance}  & 25.3 &  8.5 & 30.2 & 10.2  \\
        \midrule
        SSG~\cite{fu2019self}           & 31.6 & 13.2 & 32.2 & 13.3  \\
        \method-SSG (ours)     & 45.5 & 19.1 & 43.3 & 17.9  \\ 
        
        \midrule
        MMT~\cite{ge2020mutual}         & 54.4 & 26.6 & 58.2 & 29.3  \\
        MMT (DBSCAN)                    & 51.6 & 26.6 & 59.0 & 32.0  \\
        \method-MMT (ours) & \textbf{61.7}& \textbf{34.6} & \textbf{66.9} & \textbf{38.3} \\
        \bottomrule
        \end{tabular}
\end{table}

\subsection{Ablation Study}
\label{subsec:ablation}
In this section, we first perform a study to evaluate the impact of the value of $\mu$. Secondly, we demonstrate the interest of the conditional strategy, versus its non-conditional counterpart. Thirdly,  we study the evolution of the mutual information between ground-truth camera indexes and pseudo-labels using MTT (DBSCAN), thus providing some insights on the quality of the pseudo-labels and the impact of the conditional strategy on it. Finally, we visualize the evolution of the number of lost person identities at each training epoch, to assess the impact of the variability of the training set.


\PAR{Selection of $\mu$}
\label{subsubsec:mu}
We ablate the value $\mu$ by comparing the performance (R1 and mAP) of models trained within the range $[0.01, 1.8]$. From Tab.~\ref{tab:ablationmu}, $\mu=0.1$ (\method-MMT) and $\mu=0.05$ (\method-SSG) yield the best person re-ID performance.
\begin{table}[t]
\caption{Impact of $\mu$ in the performance of \method. When the mAP values are equal, we highlight the one corresponding to higher R1.}
    \label{tab:ablationmu}
        \centering
        \begin{tabular}{c ccccc}
        \toprule
        \multirow{2}{*}{Method} & \multirow{2}{*}{$\mu$}& \multicolumn{2}{c}{Mkt\myto\ Duke} & \multicolumn{2}{c}{Duke\myto\ Mkt}          \\
        \cmidrule{3-6} 
                & & R1                & {mAP}      & R1& mAP                         \\
        \midrule
        \multirow{6}{*}{\rotatebox[origin=c]{90}{\method-SSG}} &{0.01}  &  72.8 & 53.3 &  79.7   & 57.2   \\
        &{0.05}       &   \textbf{76.1}   &   \textbf{57.0}  &   \textbf{83.3}    &   \textbf{61.9}      \\
        &{0.1}      &     74.7 & 56.2 & 82.7& 61.1                 \\
        
        &{0.2}         &     75.3   & 56.5     &     81.8  &       60.3           \\
        
        &{0.4}        &      73.3    &    53.5        &   80.4    & 59.2          \\
        
        &{1.8}     &          7.1   &       2.9    &    39.1   &       17.1           \\
        \midrule
        \multirow{6}{*}{\rotatebox[origin=c]{90}{\method-MMT}} &{0.01} &   81.3  &  68.9 &  92.6 & 79.2\\
        &{0.05} & 82.4    &   70.3         &     93.0  & 81.3                  \\
        &{0.1}  &     \textbf{83.3} & \textbf{70.3} &\textbf{94.2} & \textbf{83.0}              \\
        
        &{0.2}      &         82.7    &      70.3    &     93.4  & 82.5                  \\
        
        &{0.4}    &        82.5     & {70.3}         &      93.8 &   82.0              \\
        
        &{1.8}   &         82.8    &    69.9    &    93.1   & 81.3                 \\
        \bottomrule
        \end{tabular}
\end{table}

\begin{table}[t]
\caption{Evaluation of the impact of the conditional strategy on SGG~\cite{fu2019self} and MMT~\cite{ge2020mutual} (using DSCAN).  When the mAP values are equal, we highlight the one corresponding to higher R1.\vspace{-3mm}}
    \label{tab:ablation}
\center
        \centering
        \begin{tabular}{cccccccc}
        \toprule
        \multicolumn{4}{c}{\multirow{2}{*}{Method}}   & \multicolumn{2}{c}{Mkt\myto\ Duke}         & \multicolumn{2}{c}{Duke\myto\ Mkt}          \\
        \cmidrule{5-8} 
         & & &       & R1                & \multicolumn{1}{c}{mAP}      & R1& mAP                         \\
        \midrule
        \multicolumn{4}{c}{SSG~\cite{fu2019self}}       &73.0                 &53.4                 &80.0     &58.3\\
        \multicolumn{4}{c}{SSG+Adv.}     & 75.4                & 56.4      & \textbf{83.8}       & \textbf{62.7}                 \\ 
        \multicolumn{4}{c}{\method-SSG}  &  \textbf{76.1}   &   \textbf{57.0}  &   83.3    &   61.9      \\ 
        \midrule
        \multicolumn{4}{c}{MMT (DBSCAN)}         & 80.2                 & 67.2               & 91.7     & 79.3              \\
        \multicolumn{4}{c}{MMT+Adv.}        & 82.6                 & 70.3                 & 93.6         & 82.2                 \\
        \multicolumn{4}{c}{\method-MMT}         & \textbf{83.3}                & \textbf{70.3}                & \textbf{94.2}                 & \textbf{83.0}                 \\
        \bottomrule
        \end{tabular}
\end{table}

\PAR{Is conditional necessary?}
From Table~\ref{tab:ablation}, we show that the camera adversarial network can help the person re-ID networks trained with clustering-based unsupervised methods better capture the person identity features: \method\ and adding a simple adversarial discriminator (+Adv.) significantly outperform the baseline methods in all settings. This is due to the combination of the camera adversarial network with unsupervised clustering-based methods. By doing so, the camera dependency is removed from the features of each person thus increasing the quality of the overall clustering. However, because of the negative transfer effect, the camera adversarial network cannot fully exploit the camera information while discarding the person ID information. For this reason, the proposed method \method~improves the capacity of the camera adversarial network over the simple adversarial strategy. In summary, we demonstrate that the camera adversarial network can help improve the results of unsupervised clustering-based person re-ID. Moreover, the proposed \method~further improves the results by removing the link between camera and IDs.

\PAR{Removing camera information}
\begin{figure}[t]
    \centering
\subfigure[ Mkt\myto\ Duke]{\includegraphics[width=0.45\textwidth]{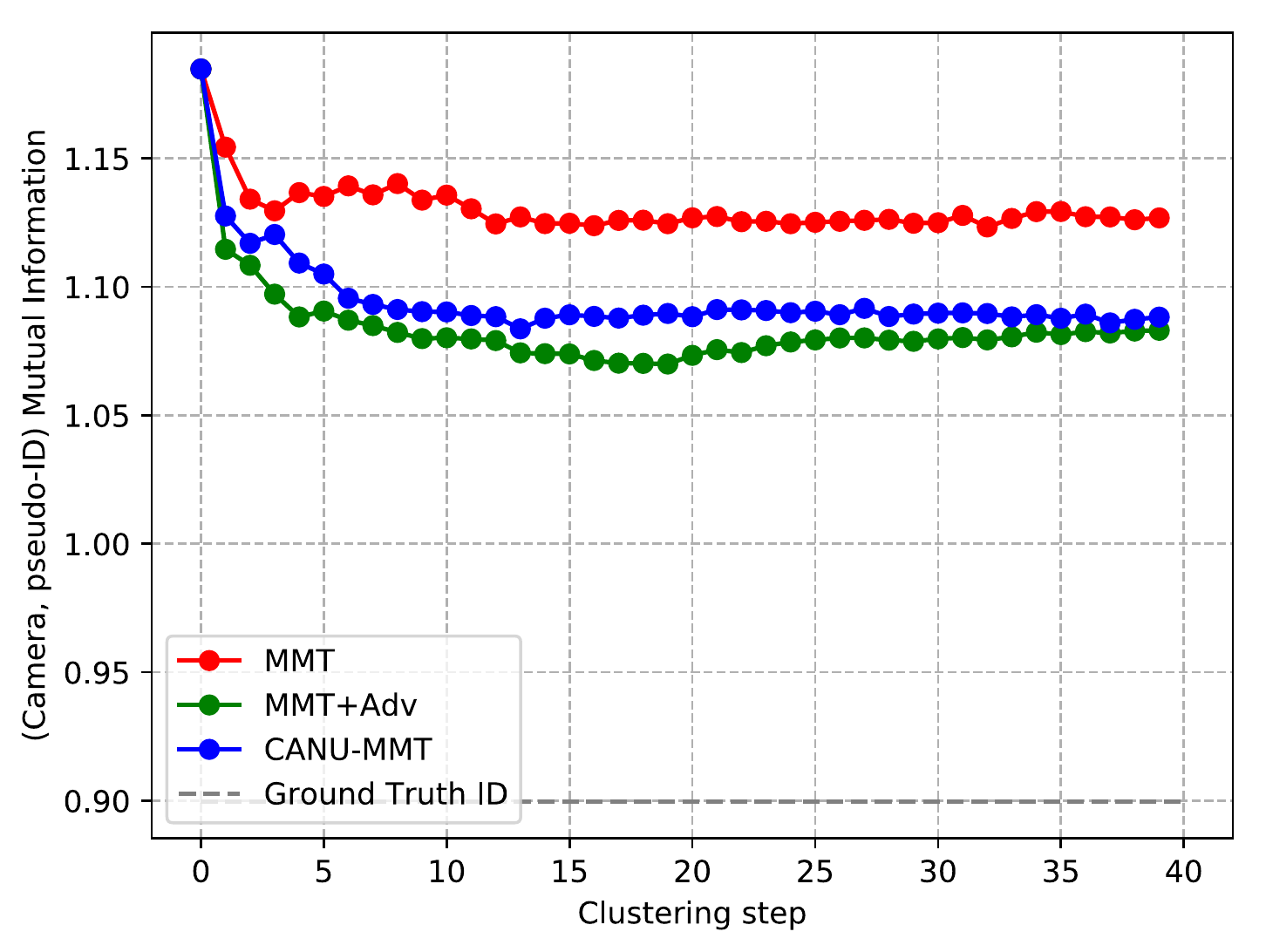}}

\subfigure[Duke\myto\ Mkt]{\includegraphics[width=0.45\textwidth]{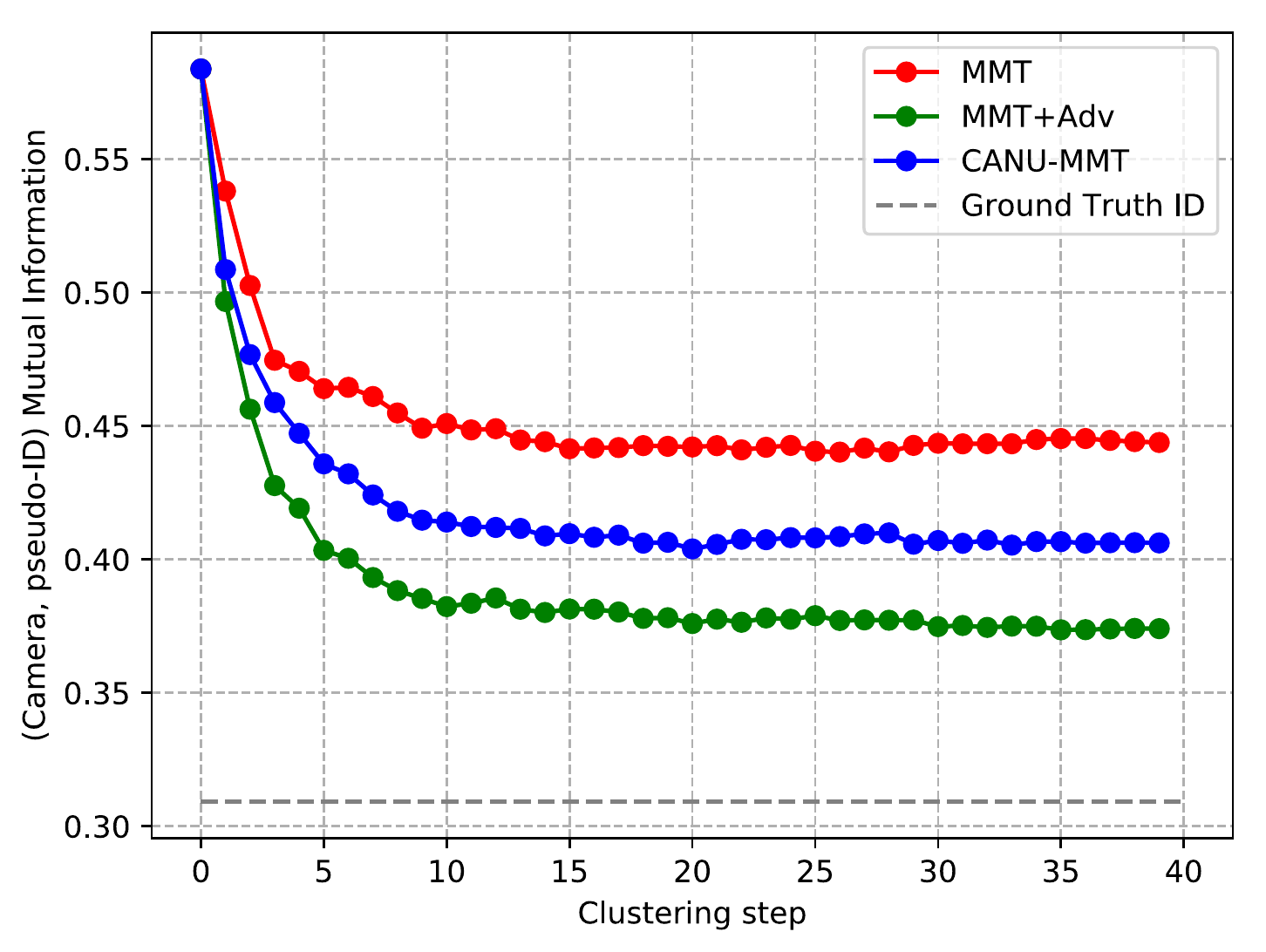}}
\vspace*{-3mm}
\caption{\label{fig:mutual_info} Mutual information between pseudo labels and camera index evolution for the MMT setting. Ground-truth ID comparison is displayed in dashed lines for both datasets.}
\vspace*{-3mm}
\end{figure}
Table~\ref{tab:ablation} demonstrates that removing camera information is globally positive, but that can also be harmful if it is not done with care.  In this section, we further demonstrate that the proposed adversarial strategies actually reduce the camera dependency in clustering results and present some insights on why the conditional strategy is better than the plain adversarial network. To do so, we plot the mutual information between the pseudo-labels provided by DBSCAN, and the fixed camera index information, at each clustering stage (i.e.\,training epoch) in Fig.~\ref{fig:mutual_info}. Intuitively, the mutual information between two variables is a measure of mutual dependence between them: the higher it is, the more predictable one is from knowing the other. We report the results for MMT on Duke\myto\ Mkt and Mkt\myto\ Duke, \method-MMT and the simple adversarial strategy. We observe that the mutual information is systematically decreasing with the training, even for plain MMT. Both adversarial strategies significantly outperform plain MMT at reducing the camera-pseudo-ID dependency, \method-MMT being slightly less effective than MMT+Adv. This is consistent with our theoretical framework, since matching ID-conditioned camera distribution in $\phi$ does not account for the ID-Camera dependency, and thus is less effective in terms of camera dependency, but preserves identity information, see Table~\ref{tab:ablation}. We also observe that there is a significant gap between the target mutual information (i.e.\, measured between ground truth ID and camera index) for all methods, which exhibits the performance gap between supervised and unsupervised person re-ID methods.

\begin{figure}[t]
\centering
\includegraphics[width=0.45\textwidth]{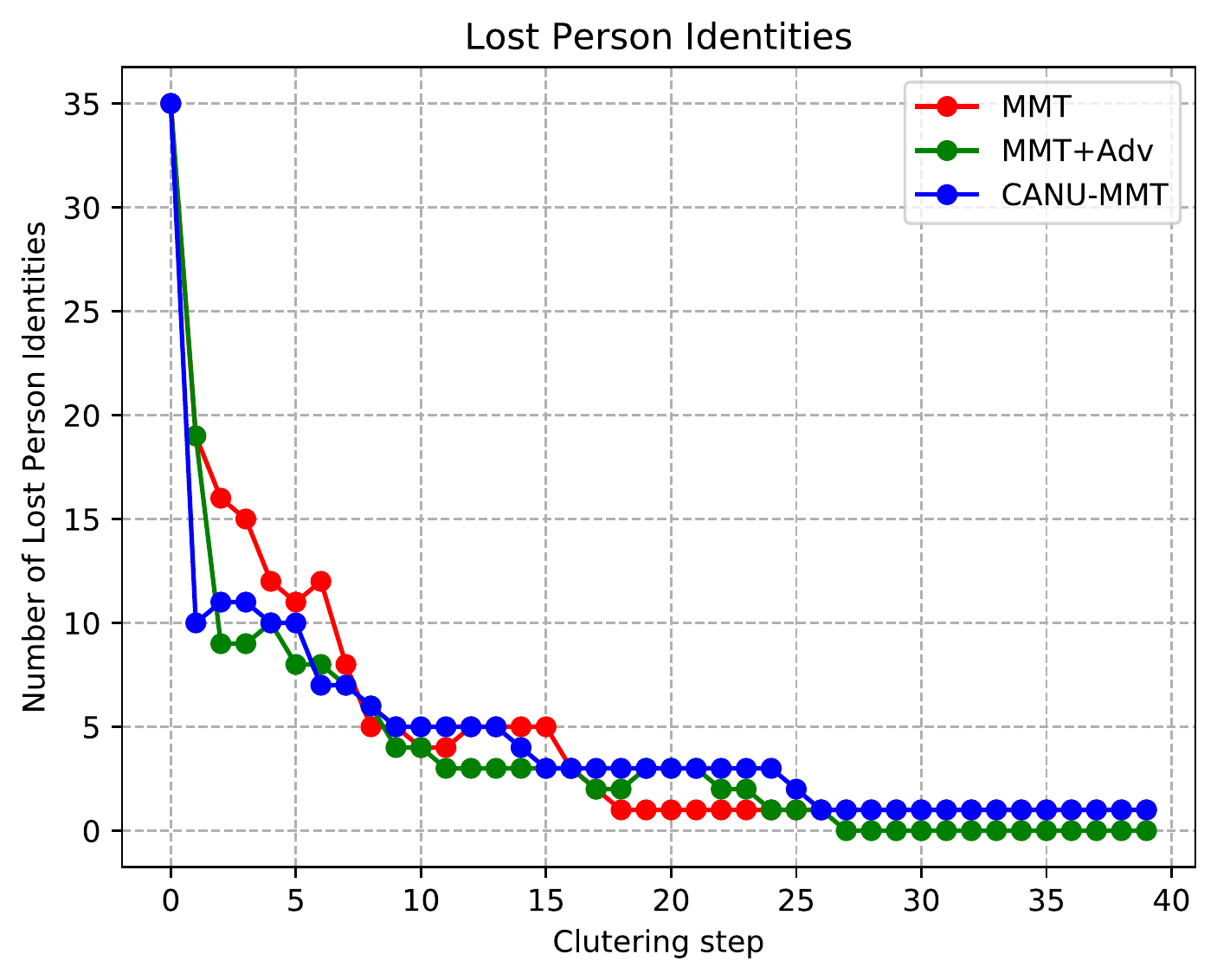}

    \vspace*{-5mm}
    \caption{Evolution of the number of lost person IDs during training using MMT on Duke\myto\ Mkt.}
    \vspace*{-4mm}
    \label{fig:numberclusters}
\end{figure}

\PAR{Evolution of the number of lost IDs}
Since we train the target dataset using unsupervised techniques, we do not use the ground-truth labels in the target dataset during training. Instead, we make use of the pseudo labels provided by DBSCAN.
DBSCAN discards the outliers i.e.\, features that are not closed to others. It is natural to wonder how many identities are ``lost'' at every iteration. We here visualize the number of lost ID (all those that are not present in a training epoch) after each clustering step. We plot the evolution of this number with the training epoch for MMT, MMT+Adv.\ and \method-MMT on Duke\myto\ Mkt in Fig.~\ref{fig:numberclusters}. The dual experiment, i.e.\ on Mkt\myto\ Duke revealed that no ID was lost by any method. In Fig.~\ref{fig:numberclusters}, we first observe that the loss of person identities decreases with the clustering steps. It means that the feature extractor provides more and more precise features representing person identities. Secondly, the use of camera adversarial training can reduce the loss of person identities in the clustering algorithm, which reflects the benefit of camera adversarial networks to the clustering algorithm and thus to the unsupervised person re-ID task.

\section{Conclusion}
\label{sec:conclusions}
In this paper, we demonstrate the benefit of unifying adversarial learning with current unsupervised clustering-based person re-identification methods. We propose to condition the adversarial learning with the cluster centroids, being these representations independent of the number of clusters and invariant to cluster index permutations. The proposed strategy boosts existing clustering-based unsupervised person re-ID baselines and sets the new state-of-the-art performance in four different unsupervised person re-ID experimental settings. We believe that the proposed method \method~was a missing component in training unsupervised person re-identification networks and we hope that our work can give insight to this direction in the person re-identification domain.






%



{\small
\bibliographystyle{ieeetran}
\bibliography{main}

\begin{thebibliography}{10}
\providecommand{\url}[1]{#1}
\csname url@samestyle\endcsname
\providecommand{\newblock}{\relax}
\providecommand{\bibinfo}[2]{#2}
\providecommand{\BIBentrySTDinterwordspacing}{\spaceskip=0pt\relax}
\providecommand{\BIBentryALTinterwordstretchfactor}{4}
\providecommand{\BIBentryALTinterwordspacing}{\spaceskip=\fontdimen2\font plus
\BIBentryALTinterwordstretchfactor\fontdimen3\font minus
  \fontdimen4\font\relax}
\providecommand{\BIBforeignlanguage}[2]{{%
\expandafter\ifx\csname l@#1\endcsname\relax
\typeout{** WARNING: IEEEtran.bst: No hyphenation pattern has been}%
\typeout{** loaded for the language `#1'. Using the pattern for}%
\typeout{** the default language instead.}%
\else
\language=\csname l@#1\endcsname
\fi
#2}}
\providecommand{\BIBdecl}{\relax}
\BIBdecl

\bibitem{10.1109/ICPR.2014.16}
D.~Yi, Z.~Lei, S.~Liao, and S.~Z. Li, ``Deep metric learning for person
  re-identification,'' in \emph{ICPR}, 2014.

\bibitem{385d6453d2f04126994b710df0df5c96}
T.~Matsukawa and E.~Suzuki, ``Person re-identification using cnn features
  learned from combination of attributes,'' in \emph{ICPR}, 2017.

\bibitem{8545620}
N.~{Jiang}, J.~{Liu}, C.~{Sun}, Y.~{Wang}, Z.~{Zhou}, and W.~{Wu},
  ``Orientation-guided similarity learning for person re-identification,'' in
  \emph{ICPR}, 2018.

\bibitem{DUKEMTMC}
E.~Ristani, F.~Solera, R.~Zou, R.~Cucchiara, and C.~Tomasi, ``Performance
  measures and a data set for multi-target, multi-camera tracking,'' in
  \emph{ECCV Workshops}, 2016.

\bibitem{Market}
L.~Zheng, L.~Shen, L.~Tian, S.~Wang, J.~Wang, and Q.~Tian, ``Scalable person
  re-identification: A benchmark,'' in \emph{IEEE ICCV}, 2015.

\bibitem{deng2018image}
W.~Deng, L.~Zheng, Q.~Ye, G.~Kang, Y.~Yang, and J.~Jiao, ``Image-image domain
  adaptation with preserved self-similarity and domain-dissimilarity for person
  re-identification,'' in \emph{IEEE CVPR}, 2018.

\bibitem{10.1145/3243316}
H.~Fan, L.~Zheng, C.~Yan, and Y.~Yang, ``Unsupervised person re-identification:
  Clustering and fine-tuning,'' \emph{ACM Trans. Multimedia Comput. Commun.
  Appl.}, 2018.

\bibitem{fu2019self}
Y.~Fu, Y.~Wei, G.~Wang, Y.~Zhou, H.~Shi, and T.~S. Huang, ``Self-similarity
  grouping: A simple unsupervised cross domain adaptation approach for person
  re-identification,'' in \emph{IEEE ICCV}, 2019.

\bibitem{ge2020mutual}
Y.~Ge, D.~Chen, and H.~Li, ``Mutual mean-teaching: Pseudo label refinery for
  unsupervised domain adaptation on person re-identification,'' \emph{ICLR},
  2020.

\bibitem{Tzeng}
E.~Tzeng, J.~Hoffman, K.~Saenko, and T.~Darrell, ``Adversarial discriminative
  domain adaptation,'' in \emph{IEEE CVPR}, 2017.

\bibitem{ganin2016domain}
Y.~Ganin, E.~Ustinova, H.~Ajakan, P.~Germain, H.~Larochelle, F.~Laviolette,
  M.~Marchand, and V.~Lempitsky, ``Domain-adversarial training of neural
  networks,'' \emph{JMLR}, 2016.

\bibitem{domain_sep_net}
K.~Bousmalis, G.~Trigeorgis, N.~Silberman, D.~Krishnan, and D.~Erhan, ``Domain
  separation networks,'' in \emph{NIPS}, 2016.

\bibitem{CamSty}
Z.~Zhong, L.~Zheng, Z.~Zheng, S.~Li, and Y.~Yang, ``Camera style adaptation for
  person re-identification,'' in \emph{IEEE CVPR}, 2018.

\bibitem{DUKE_REID}
Z.~Zheng, L.~Zheng, and Y.~Yang, ``Unlabeled samples generated by gan improve
  the person re-identification baseline in vitro,'' in \emph{IEEE ICCV}, 2017.

\bibitem{lin2019bottom}
Y.~Lin, X.~Dong, L.~Zheng, Y.~Yan, and Y.~Yang, ``A bottom-up clustering
  approach to unsupervised person re-identification,'' in \emph{AAAI}, 2019.

\bibitem{zhang2019self}
X.~Zhang, J.~Cao, C.~Shen, and M.~You, ``Self-training with progressive
  augmentation for unsupervised cross-domain person re-identification,'' in
  \emph{IEEE CVPR}, 2019.

\bibitem{zhong2018generalizing}
Z.~Zhong, L.~Zheng, S.~Li, and Y.~Yang, ``Generalizing a person retrieval model
  hetero-and homogeneously,'' in \emph{ECCV}, 2018.

\bibitem{chang2019disjoint}
X.~Chang, Y.~Yang, T.~Xiang, and T.~M. Hospedales, ``Disjoint label space
  transfer learning with common factorised space,'' in \emph{AAAI}, 2019.

\bibitem{qi2019novel}
L.~Qi, L.~Wang, J.~Huo, L.~Zhou, Y.~Shi, and Y.~Gao, ``A novel unsupervised
  camera-aware domain adaptation framework for person re-identification,'' in
  \emph{IEEE CVPR}, 2019.

\bibitem{song2020unsupervised}
L.~Song, C.~Wang, L.~Zhang, B.~Du, Q.~Zhang, C.~Huang, and X.~Wang,
  ``Unsupervised domain adaptive re-identification: Theory and practice,''
  \emph{Pattern Recognition}, 2020.

\bibitem{zhong2019invariance}
Z.~Zhong, L.~Zheng, Z.~Luo, S.~Li, and Y.~Yang, ``Invariance matters: Exemplar
  memory for domain adaptive person re-identification,'' in \emph{IEEE CVPR},
  2019.

\bibitem{torrey.handbook09}
L.~Torrey and J.~Shavlik, ``Transfer learning,'' in \emph{Handbook of Research
  on Machine Learning Applications}, E.~Soria, J.~Martin, R.~Magdalena,
  M.~Martinez, and A.~Serrano, Eds., 2009.

\bibitem{ETN_2019_CVPR}
M.~L. J. W. Q.~Y. Zhangjie~Cao, Kaichao~You, ``Learning to transfer examples
  for partial domain adaptation,'' in \emph{IEEE CVPR}, June 2019.

\bibitem{SAN}
Z.~Cao, M.~Long, J.~Wang, and M.~I. Jordan, ``Partial transfer learning with
  selective adversarial networks,'' in \emph{IEEE CVPR}, 2018.

\bibitem{IWAN}
J.~Zhang, Z.~Ding, W.~Li, and P.~Ogunbona, ``Importance weighted adversarial
  nets for partial domain adaptation,'' \emph{IEEE CVPR}, 2018.

\bibitem{wang2019characterizing}
Z.~Wang, Z.~Dai, B.~P{\'o}czos, and J.~Carbonell, ``Characterizing and avoiding
  negative transfer,'' in \emph{IEEE CVPR}, 2019, pp. 11\,293--11\,302.

\bibitem{STN}
Y.~Yao, Y.~Zhang, X.~Li, and Y.~Ye, ``Heterogeneous domain adaptation via soft
  transfer network,'' in \emph{ACM MM}, 2019.

\bibitem{Li_2018_ECCV}
Y.~Li, X.~Tian, M.~Gong, Y.~Liu, T.~Liu, K.~Zhang, and D.~Tao, ``Deep domain
  generalization via conditional invariant adversarial networks,'' in
  \emph{ECCV}, 2018.

\bibitem{ResNet}
K.~He, X.~Zhang, S.~Ren, and J.~Sun, ``Deep residual learning for image
  recognition,'' in \emph{IEEE CVPR}, 2016.

\bibitem{DBLPHermansBL17}
A.~Hermans, L.~Beyer, and B.~Leibe, ``In defense of the triplet loss for person
  re-identification,'' \emph{ArXiv preprint}, 2017.

\bibitem{DBSCAN}
M.~Ester, H.-P. Kriegel, J.~Sander, X.~Xu \emph{et~al.}, ``A density-based
  algorithm for discovering clusters in large spatial databases with noise.''
  in \emph{Kdd}, vol.~96, no.~34, 1996, pp. 226--231.

\bibitem{JSD}
J.~Lin, ``Divergence measures based on the shannon entropy,'' \emph{IEEE
  Transactions on Information Theory}, 1991.

\bibitem{COND_GAN}
M.~Mirza and S.~Osindero, ``Conditional generative adversarial nets,''
  \emph{arXiv preprint arXiv:1411.1784}, 2014.

\bibitem{pan2018two}
X.~Pan, P.~Luo, J.~Shi, and X.~Tang, ``Two at once: Enhancing learning and
  generalization capacities via ibn-net,'' in \emph{ECCV}, 2018.

\bibitem{Wei_2018_CVPR}
L.~Wei, S.~Zhang, W.~Gao, and Q.~Tian, ``Person transfer gan to bridge domain
  gap for person re-identification,'' in \emph{IEEE CVPR}, 2018.

\bibitem{DPM}
P.~F. Felzenszwalb, R.~B. Girshick, D.~McAllester, and D.~Ramanan, ``Object
  detection with discriminatively trained part-based models,'' \emph{IEEE
  Trans. Pattern Anal. Mach. Intell.}, 2010.

\bibitem{ioffe2015batch}
S.~Ioffe and C.~Szegedy, ``Batch normalization: Accelerating deep network
  training by reducing internal covariate shift,'' \emph{arXiv preprint}, 2015.

\bibitem{adam}
D.~P. Kingma and J.~Ba, ``Adam: A method for stochastic optimization,''
  \emph{arXiv preprint arXiv:1412.6980}, 2014.

\bibitem{wang2018transferable}
J.~Wang, X.~Zhu, S.~Gong, and W.~Li, ``Transferable joint attribute-identity
  deep learning for unsupervised person re-identification,'' in \emph{IEEE
  CVPR}, 2018, pp. 2275--2284.

\bibitem{li2018adaptation}
Y.-J. Li, F.-E. Yang, Y.-C. Liu, Y.-Y. Yeh, X.~Du, and Y.-C. Frank~Wang,
  ``Adaptation and re-identification network: An unsupervised deep transfer
  learning approach to person re-identification,'' in \emph{IEEE CVPR
  Workshops}, 2018, pp. 172--178.

\bibitem{li2019cross}
Y.-J. Li, C.-S. Lin, Y.-B. Lin, and Y.-C.~F. Wang, ``Cross-dataset person
  re-identification via unsupervised pose disentanglement and adaptation,'' in
  \emph{IEEE ICCV}, 2019.

\bibitem{han2018co}
B.~Han, Q.~Yao, X.~Yu, G.~Niu, M.~Xu, W.~Hu, I.~Tsang, and M.~Sugiyama,
  ``Co-teaching: Robust training of deep neural networks with extremely noisy
  labels,'' in \emph{NeurIPS}, 2018.

\bibitem{wei2018person}
L.~Wei, S.~Zhang, W.~Gao, and Q.~Tian, ``Person transfer gan to bridge domain
  gap for person re-identification,'' in \emph{IEEE CVPR}, 2018.

\end{thebibliography}
}

\appendices
We provide the supplementary discussion on removing camera information and on the evolution of the number of lost person identities (IDs), by reporting the results we obtained with SSG~\cite{fu2019self} using Mkt\myto\ Duke and Duke\myto\ Mkt settings. We also provide additional plots to visualize the properties of the embedding learned in different settings and by different methods.

\section{Impact of \method~on SSG}
\begin{figure*}[ht]
    \centering
\subfigure[ Mkt\myto\ Duke, based on full-body features.]{\includegraphics[width=0.3\textwidth]{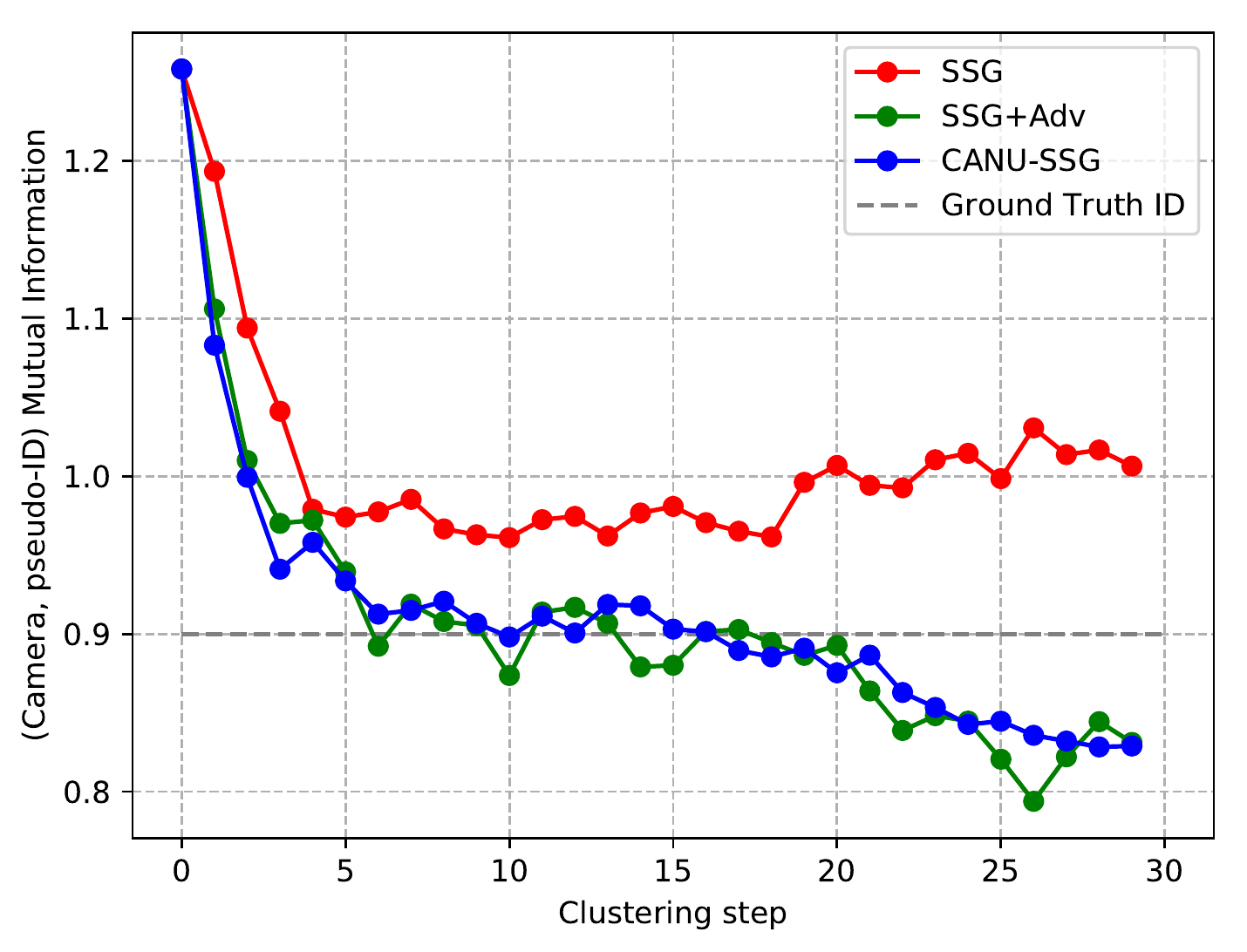}}
\subfigure[Duke\myto\ Mkt, based on full-body features.]{\includegraphics[width=0.3\textwidth]{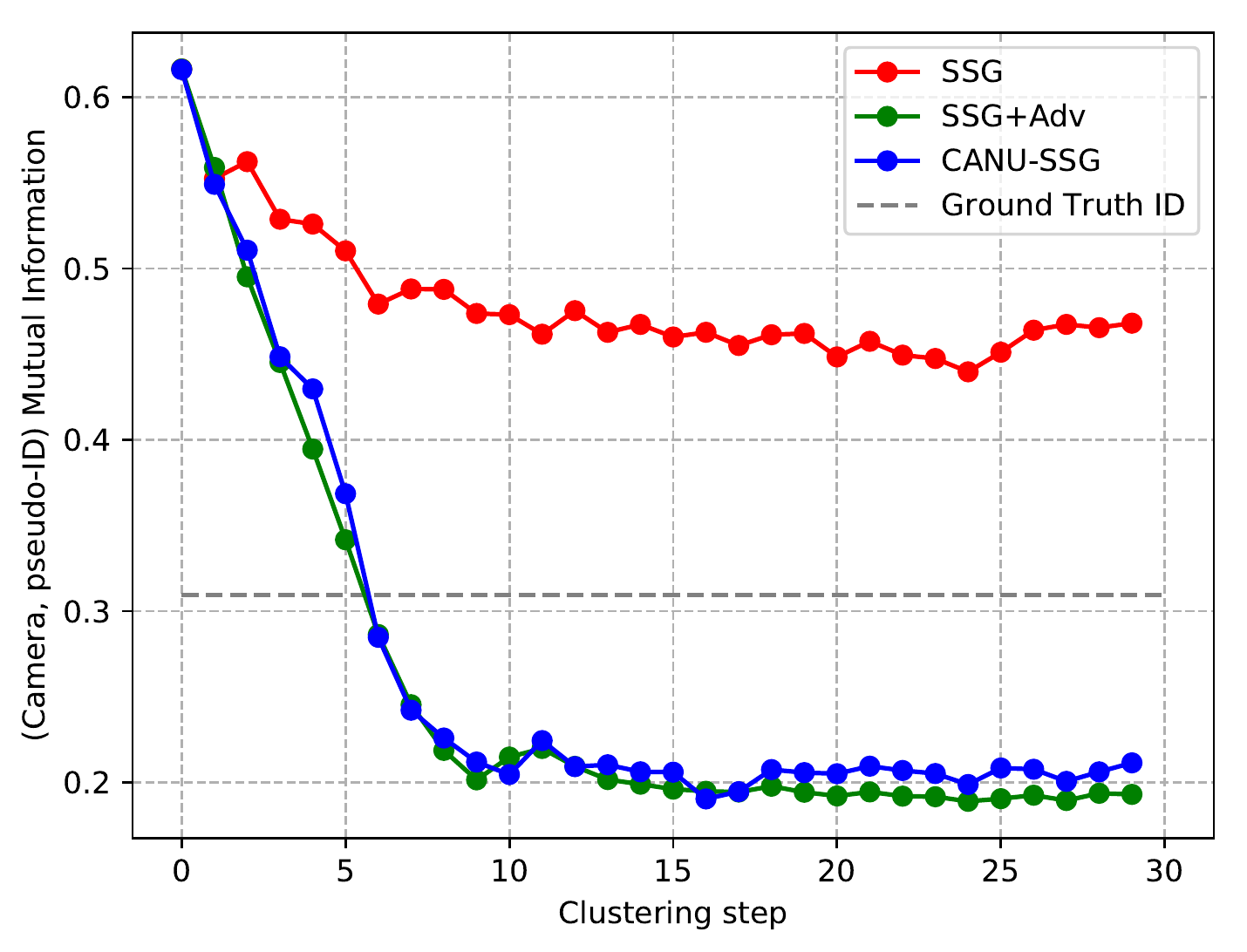}}
\subfigure[ Mkt\myto\ Duke, based on upper-body features.]{\includegraphics[width=0.3\textwidth]{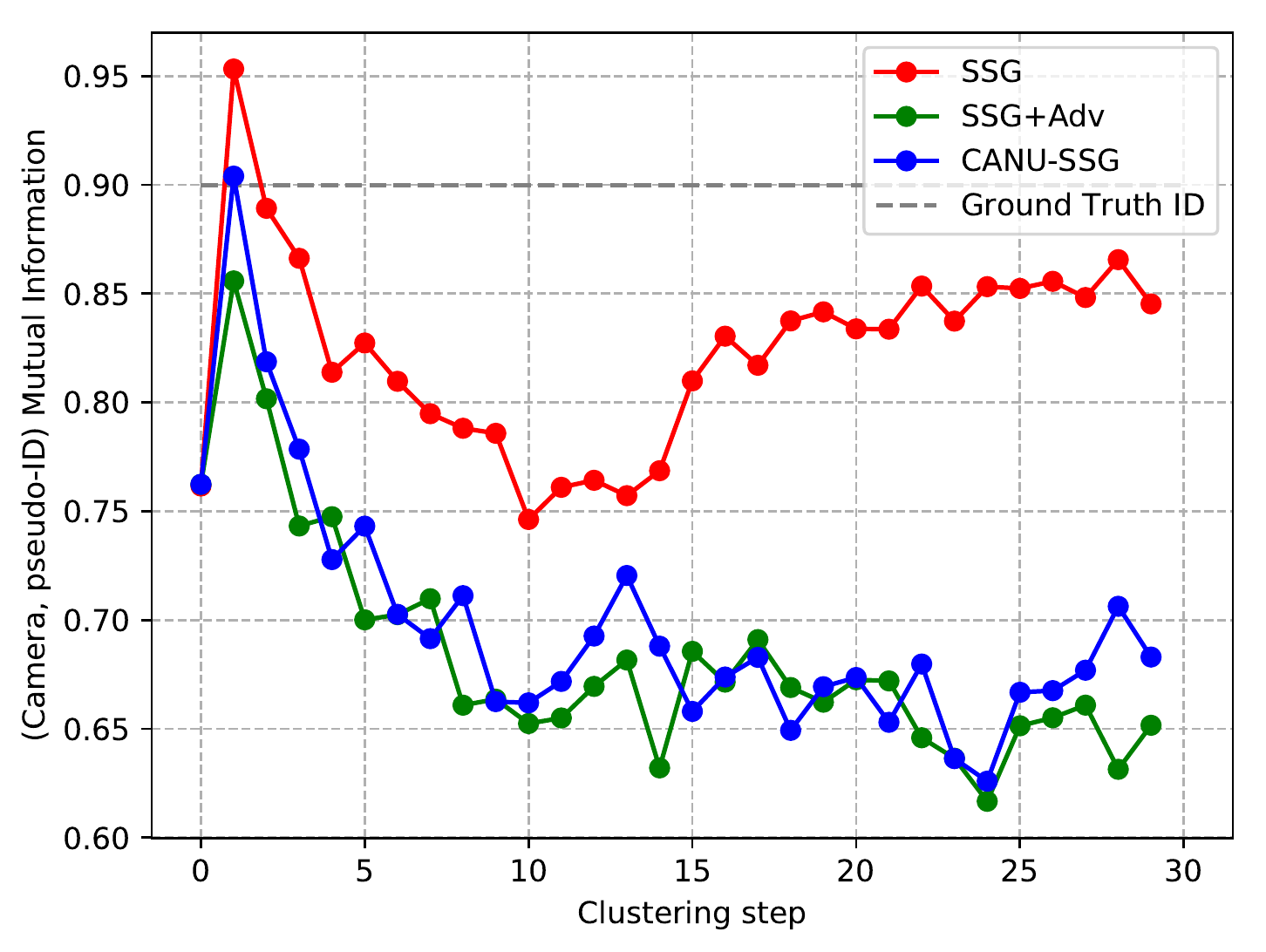}}
\subfigure[Duke\myto\ Mkt, based on upper-body features.]{\includegraphics[width=0.3\textwidth]{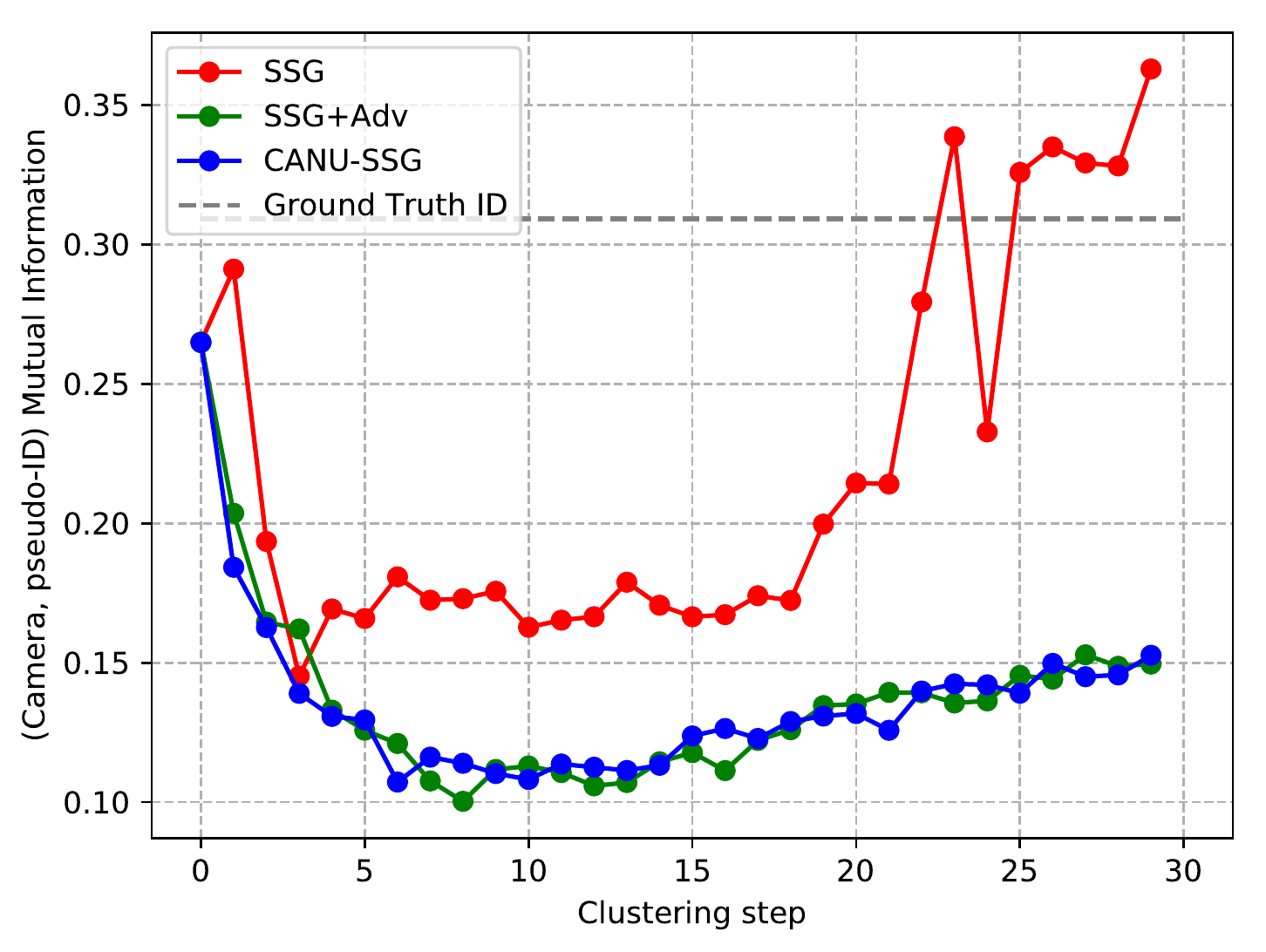}}
\subfigure[ Mkt\myto\ Duke, based on lower-body features.]{\includegraphics[width=0.3\textwidth]{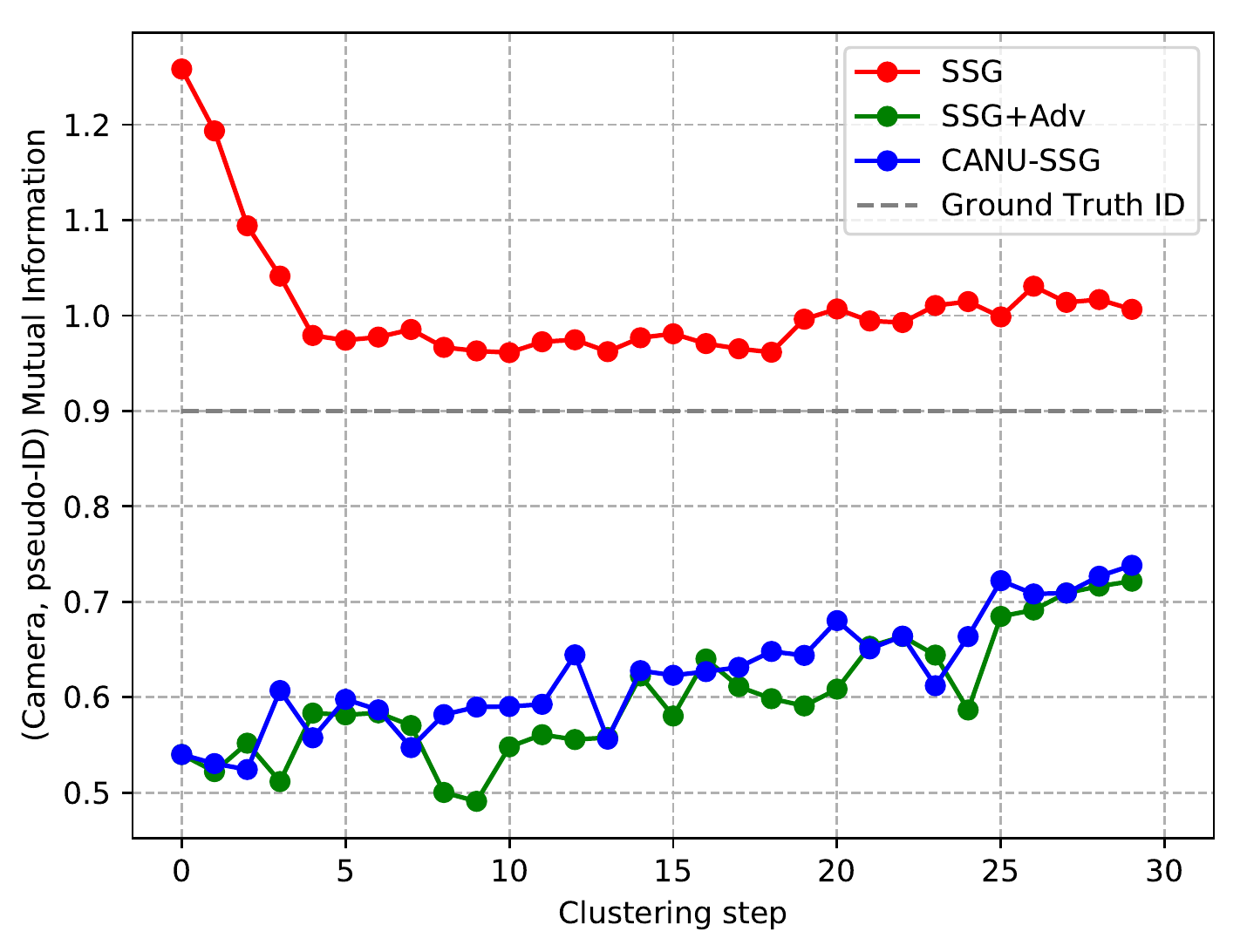}}
\subfigure[Duke\myto\ Mkt, based on lower-body features.]{\includegraphics[width=0.3\textwidth]{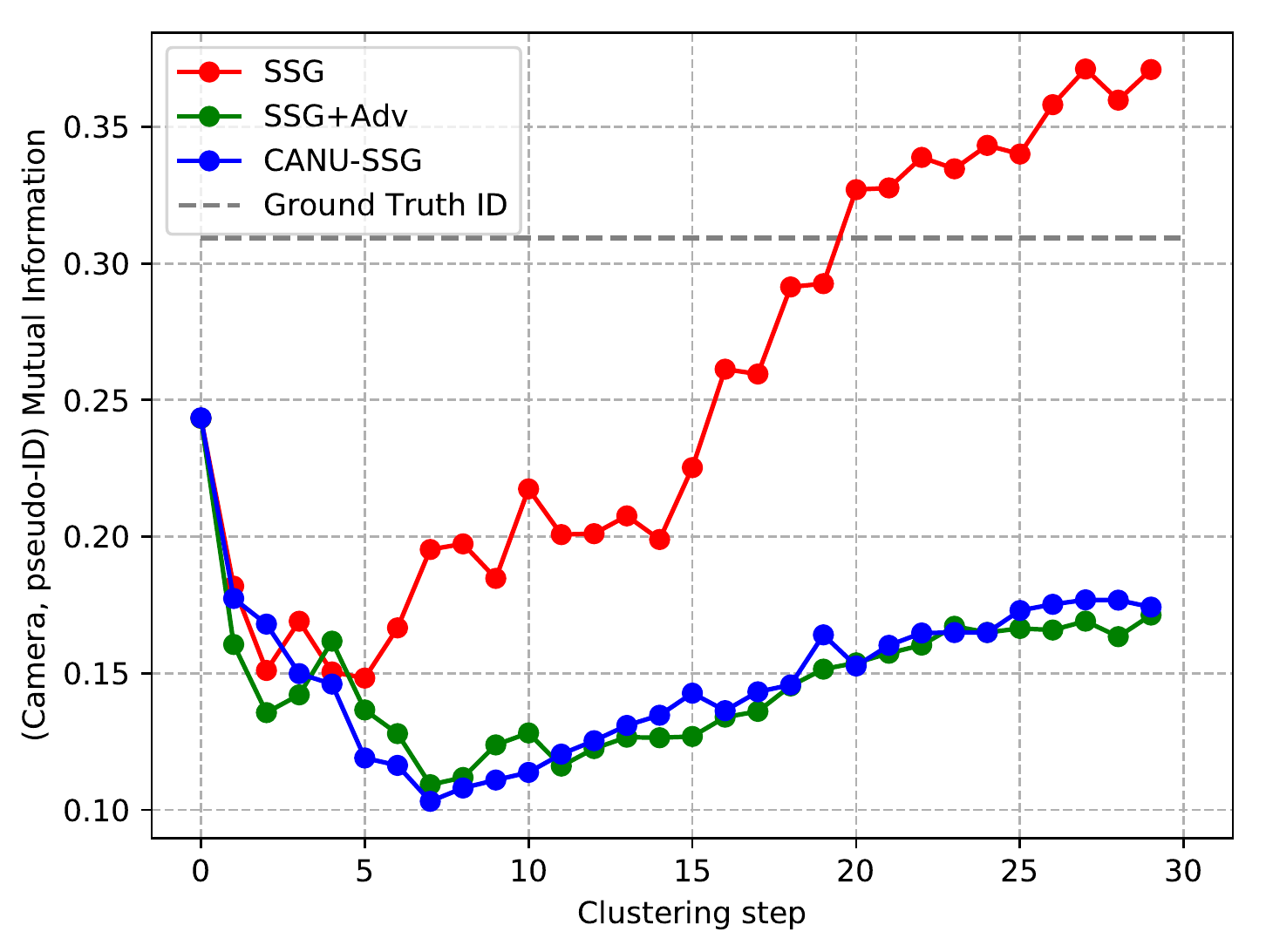}}
\vspace*{-3mm}
\caption{\label{fig:mutual_info_ssg} Mutual information between pseudo labels and camera index evolution for the SSG~\cite{fu2019self} setting. Ground-truth ID comparison is displayed in dashed lines for both datasets.}
\end{figure*}

\PAR{Removing camera information} From Fig.~\ref{fig:mutual_info_ssg}, we report the evolution of mutual information between the pseudo-labels provided by DBSCAN~\cite{DBSCAN}, and the fixed camera index information with SSG~\cite{fu2019self} over the clustering stages (i.e.\,training epochs). Similar to MMT, based on Duke\myto\ Mkt and Mkt\myto\ Duke settings, we compare SST, \method-SSG and, the simple adversarial strategy (+Adv). However, since SSG exploits different embedding spaces (i.e.\, features) and generates a clustering result for each one of them independently, we report the results for full-, upper-, and lower-body features. Moreover, We report target mutual information between ground-truth IDs and camera index. 
Similar to MMT~\cite{ge2020mutual}, we observe that plain SSG handles pseudo-ID labels significantly more dependant on camera labels than both adversarial methods, and for all representations. We also observe that \method-SSG's labels are slightly more camera dependant that SSG+Adv, even if the difference is less clear than for MMT. We also note that the camera dependency of the clustering results is much closer to the target mutual information.

\begin{figure*}[ht]
\centering
\subfigure[\# lost IDs based on full-body features.]{\includegraphics[width=0.3\textwidth]{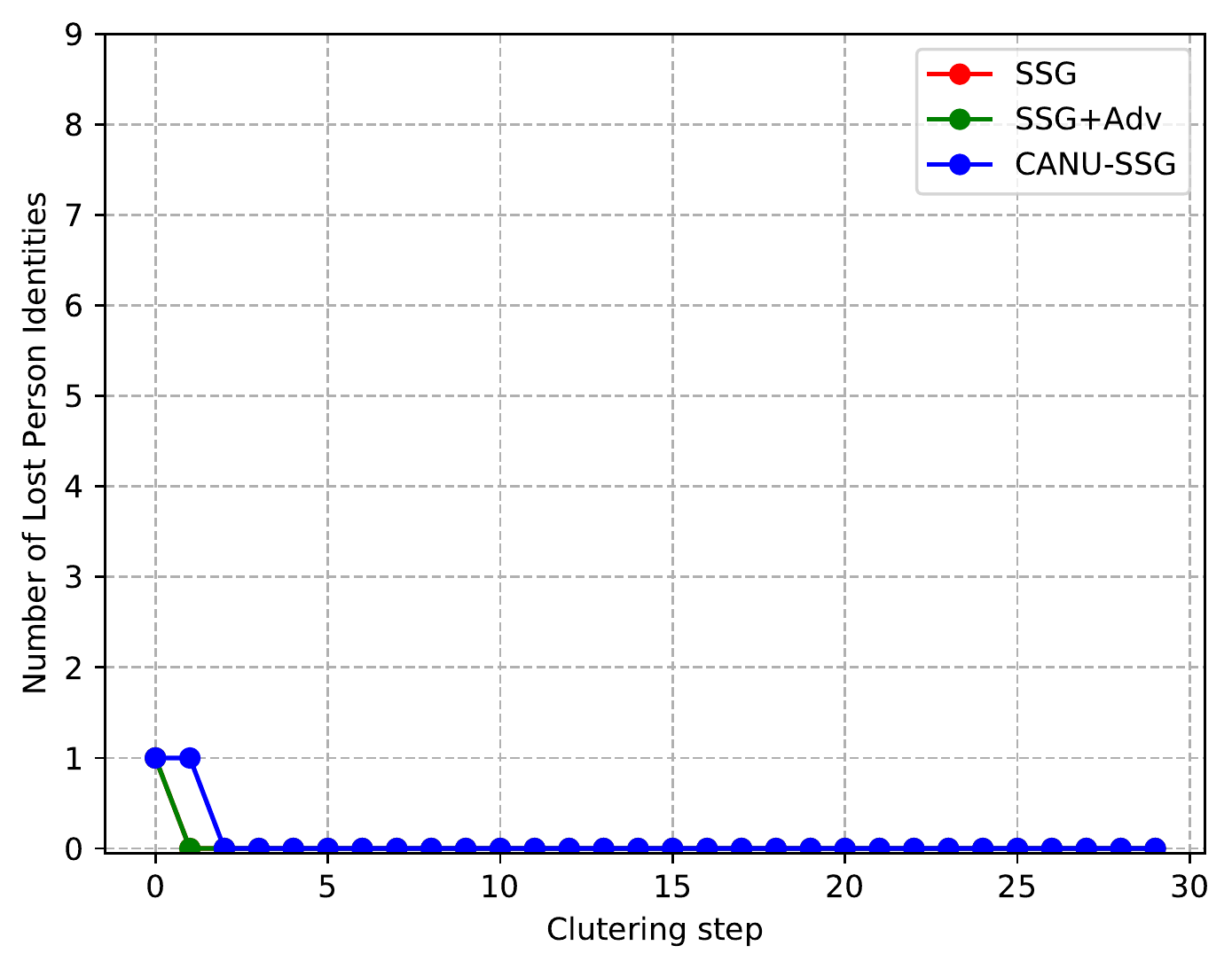}}
\subfigure[\# lost IDs based on upper-body features.]{\includegraphics[width=0.3\textwidth]{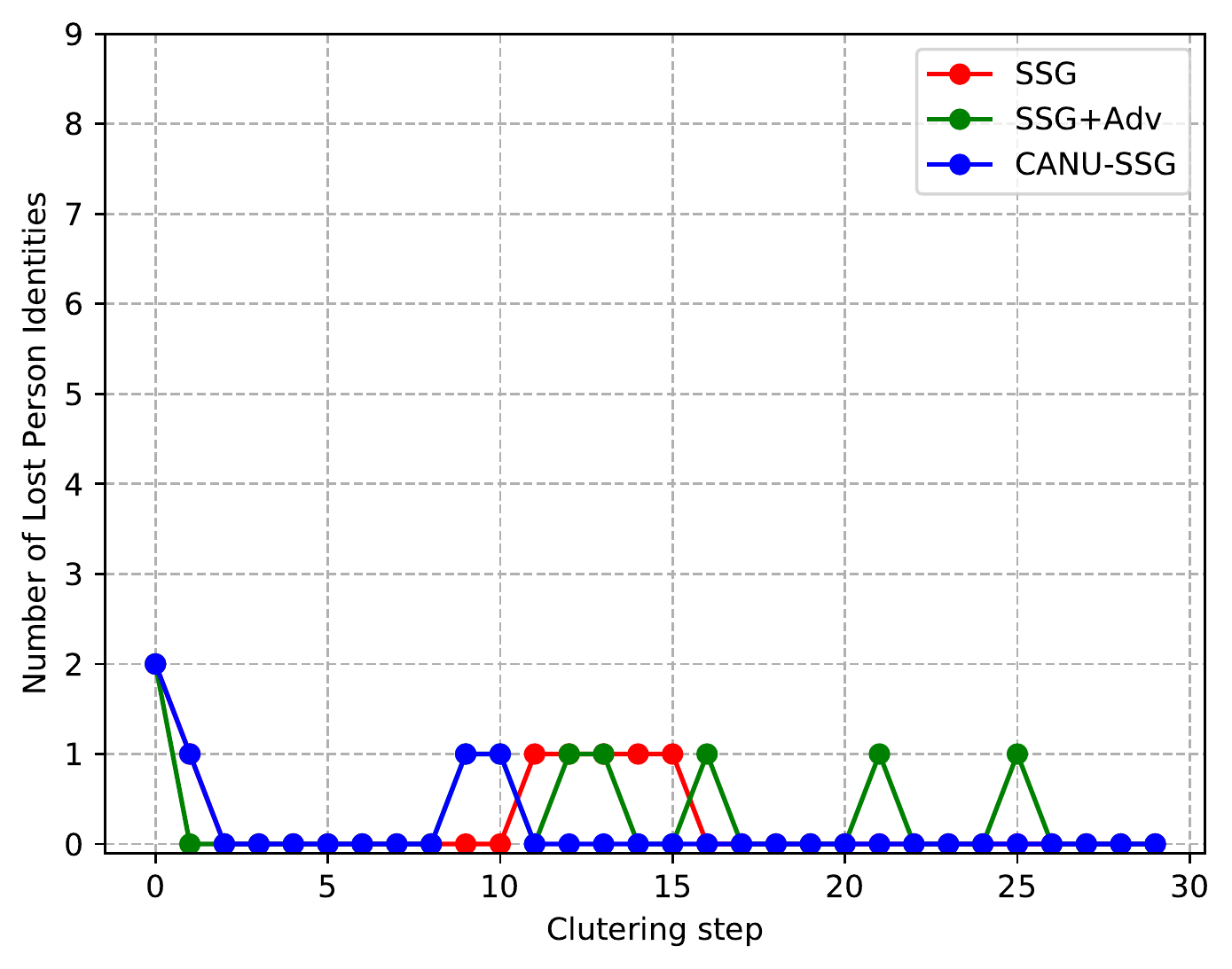}}
\subfigure[\# lost IDs based on lower-body features.]{\includegraphics[width=0.3\textwidth]{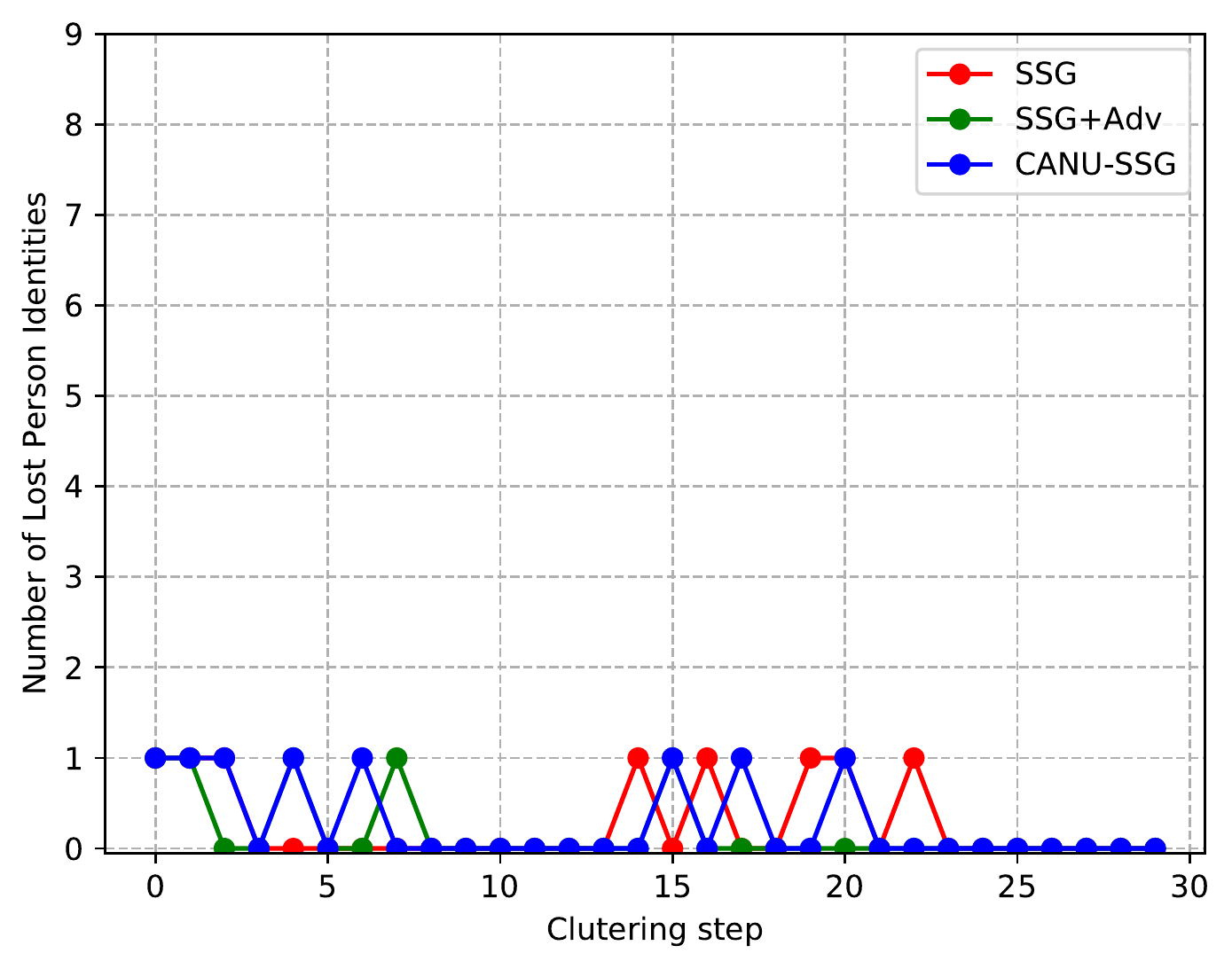}}
    \caption{Evolution of the number of lost person IDs during training using SSG~\cite{fu2019self} on Duke\myto\ Mkt. In SSG, DBSCAN clusters full- (a), upper- (b) and lower-body (c) features independently.}
    \label{fig:numberclustersSSG}
\end{figure*}
{
\PAR{Evolution of the number of lost IDs}
As discussed previously, we do not use the ground-truth labels in the target data set during training. Instead, we make use of the pseudo labels provided by DBSCAN. Since SSG~\cite{fu2019self} uses three different types of features: full-, upper- and, lower-body features, we visualize in Fig.~\ref{fig:numberclustersSSG} the number of lost identities (the ground-truth identities that are not present in pseudo labels given by DBSCAN, i.e.\, they are considered as outliers.) respectively based on the clustering results from different features.

As in MMT, no IDs are lost in the Mkt\myto\ Duke setting. For Duke\myto\ Mkt, we observe that only a very few IDs are lost ($\leq$ 2 IDs). Precisely, the number of lost person IDs from the clustering results on full-body features remains 0 except at the beginning of the first epoch (1 lost ID). Moreover, for upper-body features, All settings lose less than 2 IDs and remarkably, \method-SSG loses no IDs during most of the training epochs and it has a lower loss compared to SSG+Adv and SSG. Finally, for lower-body features, at most 1 ID is lost during the training procedure. In summary, (1) very few IDs have been lost during on SSG using DBSCAN. (2)\method-SSG has fewer losses compared to SSG+Adv and SSG. (e.g.\, Fig.~\ref{fig:numberclustersSSG} (b)).  
}

\section{Embedding visualisation}

\begin{figure*}[t]
    \centering
\subfigure[MMT, camera {\color{blue}3} and {\color{orange} 6}]{\includegraphics[width=0.3\textwidth]{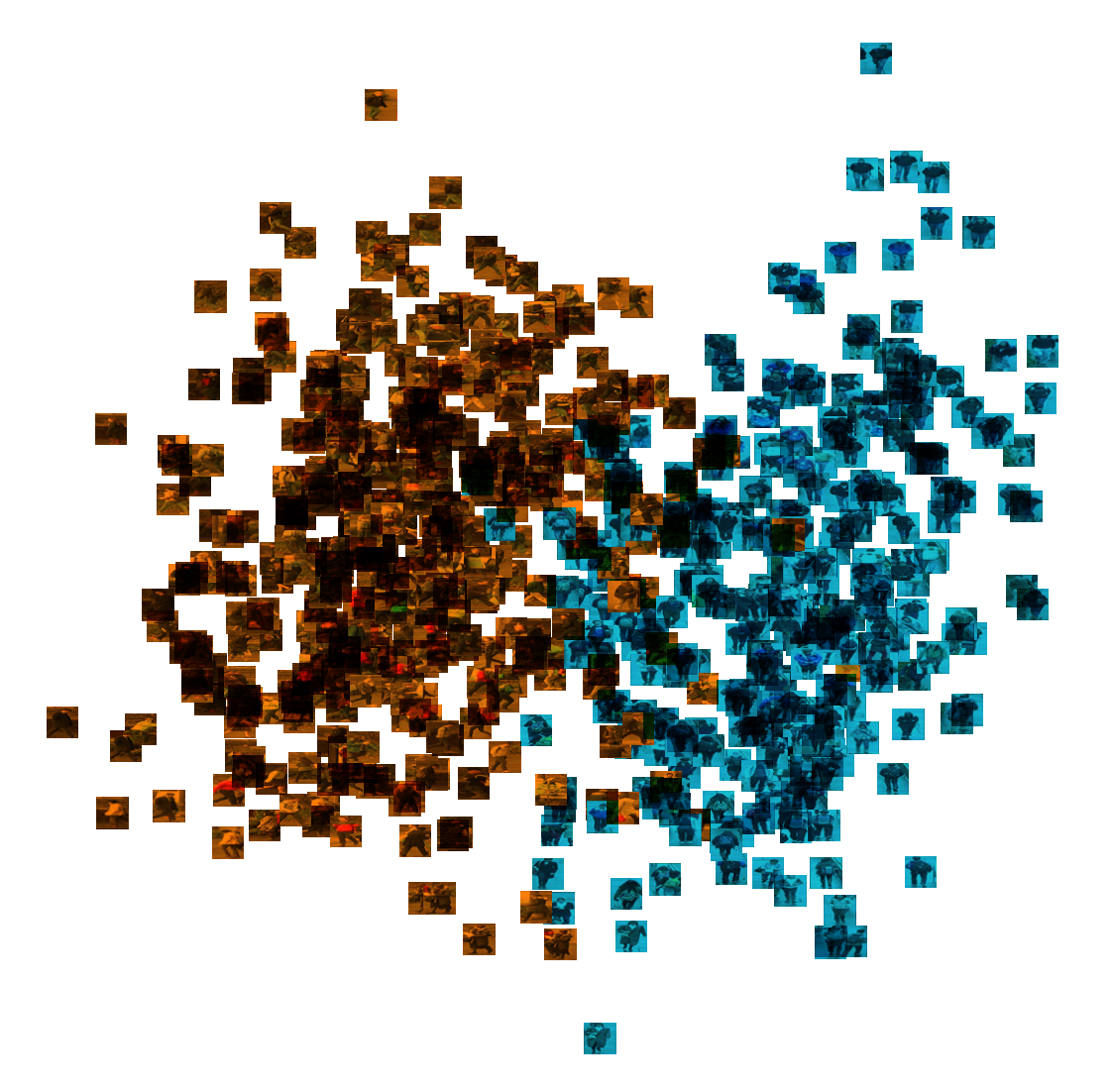}}
\subfigure[MMT+Adv, camera {\color{blue}3} and {\color{orange} 6}]{\includegraphics[width=0.3\textwidth]{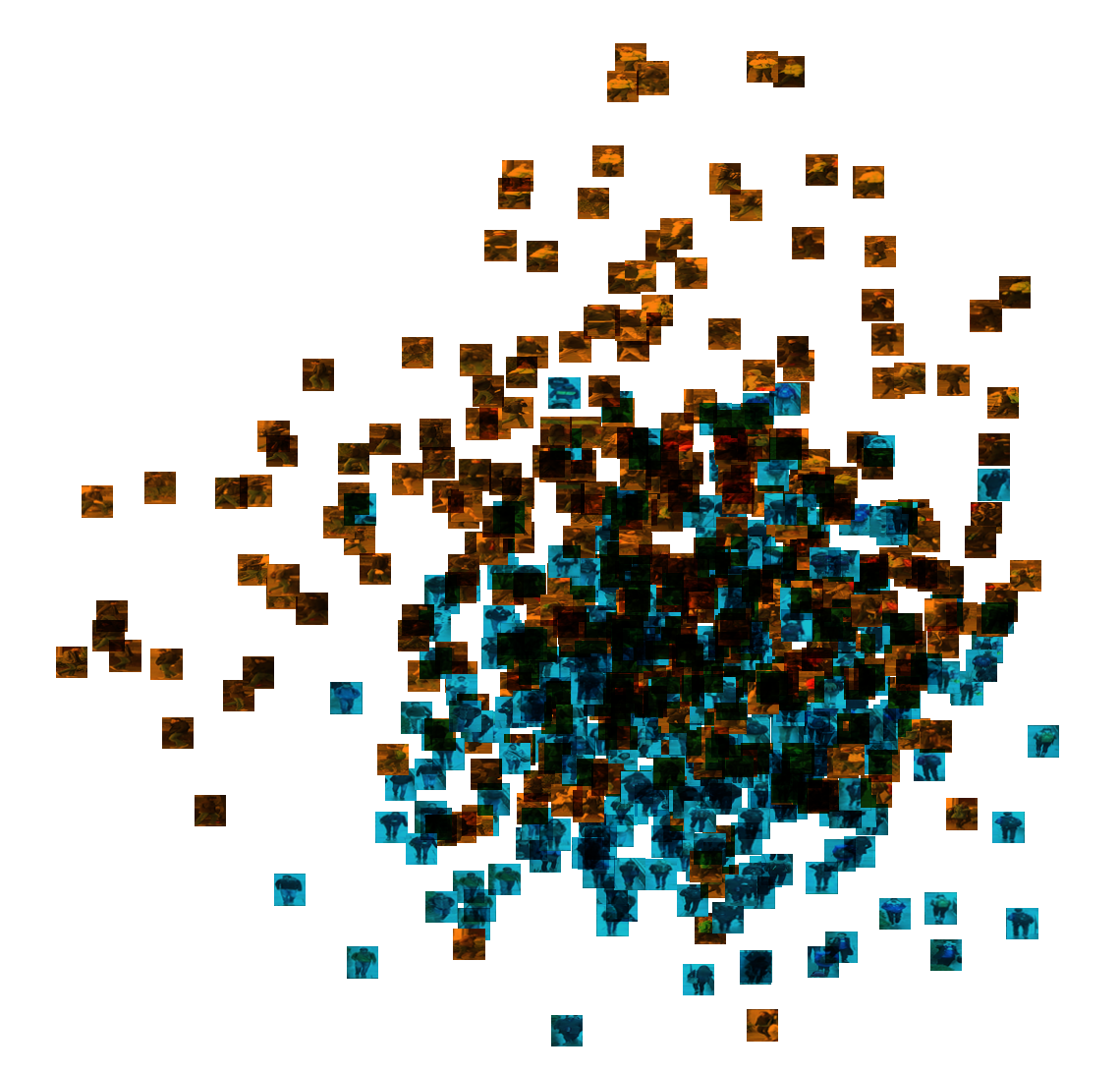}}
\subfigure[CANU-MMT, camera {\color{blue}3} and {\color{orange} 6}]{\includegraphics[width=0.3\textwidth]{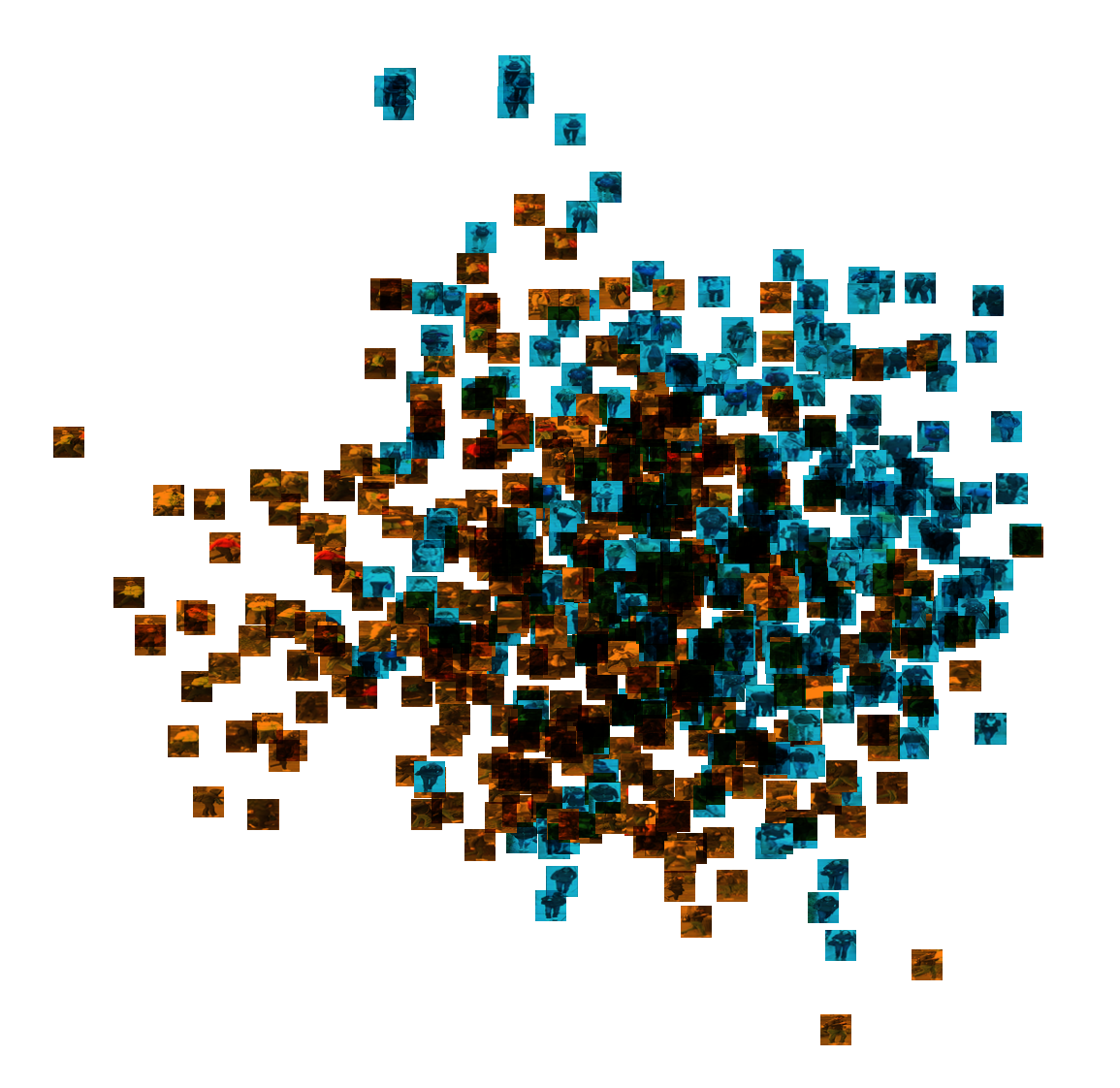}}

\subfigure[MMT, camera {\color{purple} 2} and \textcolor{green}{4}]{\includegraphics[width=0.3\textwidth]{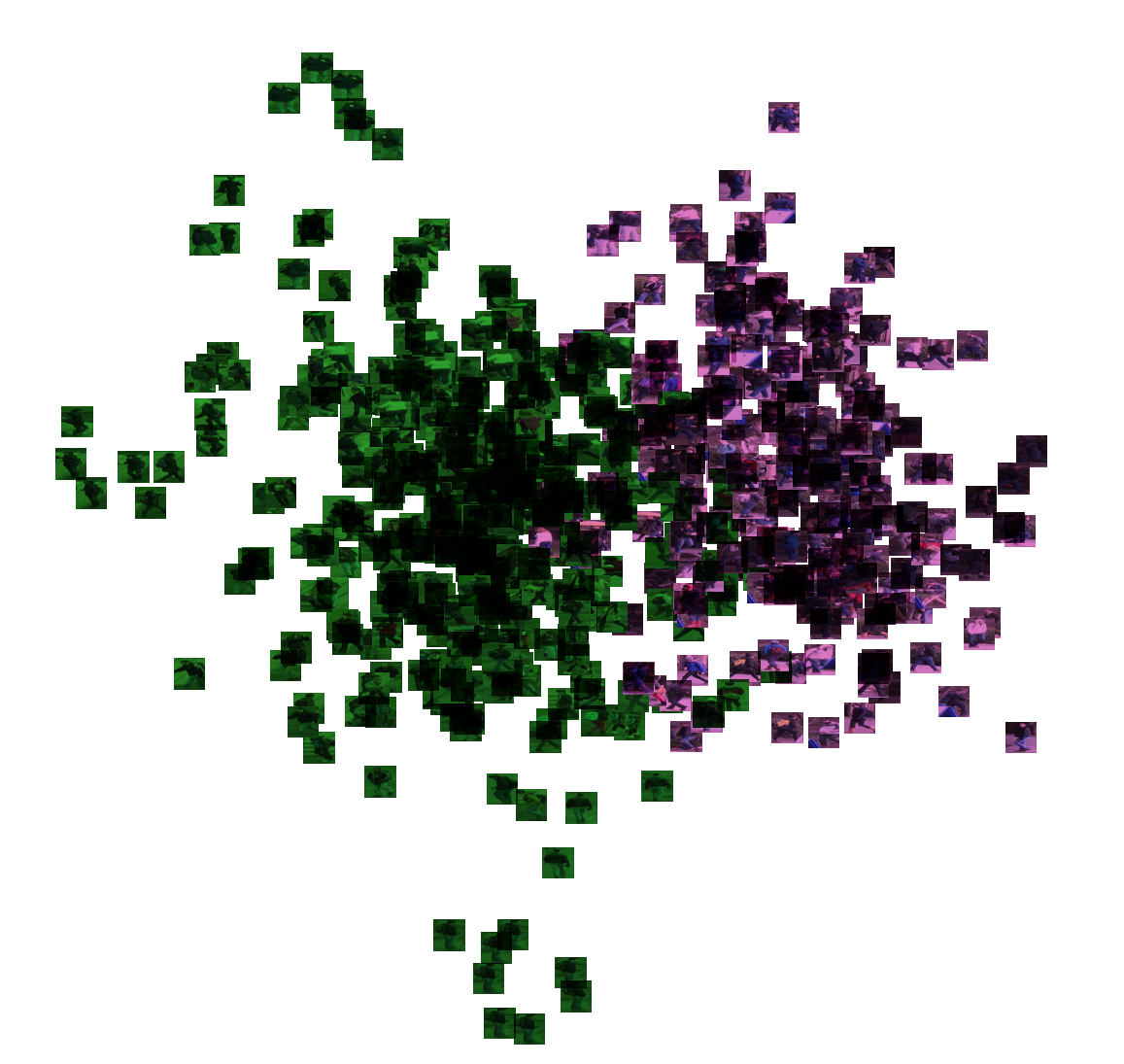}}
\subfigure[MMT+Adv, {\color{purple} 2} and {\color{green} 4}]{\includegraphics[width=0.3\textwidth]{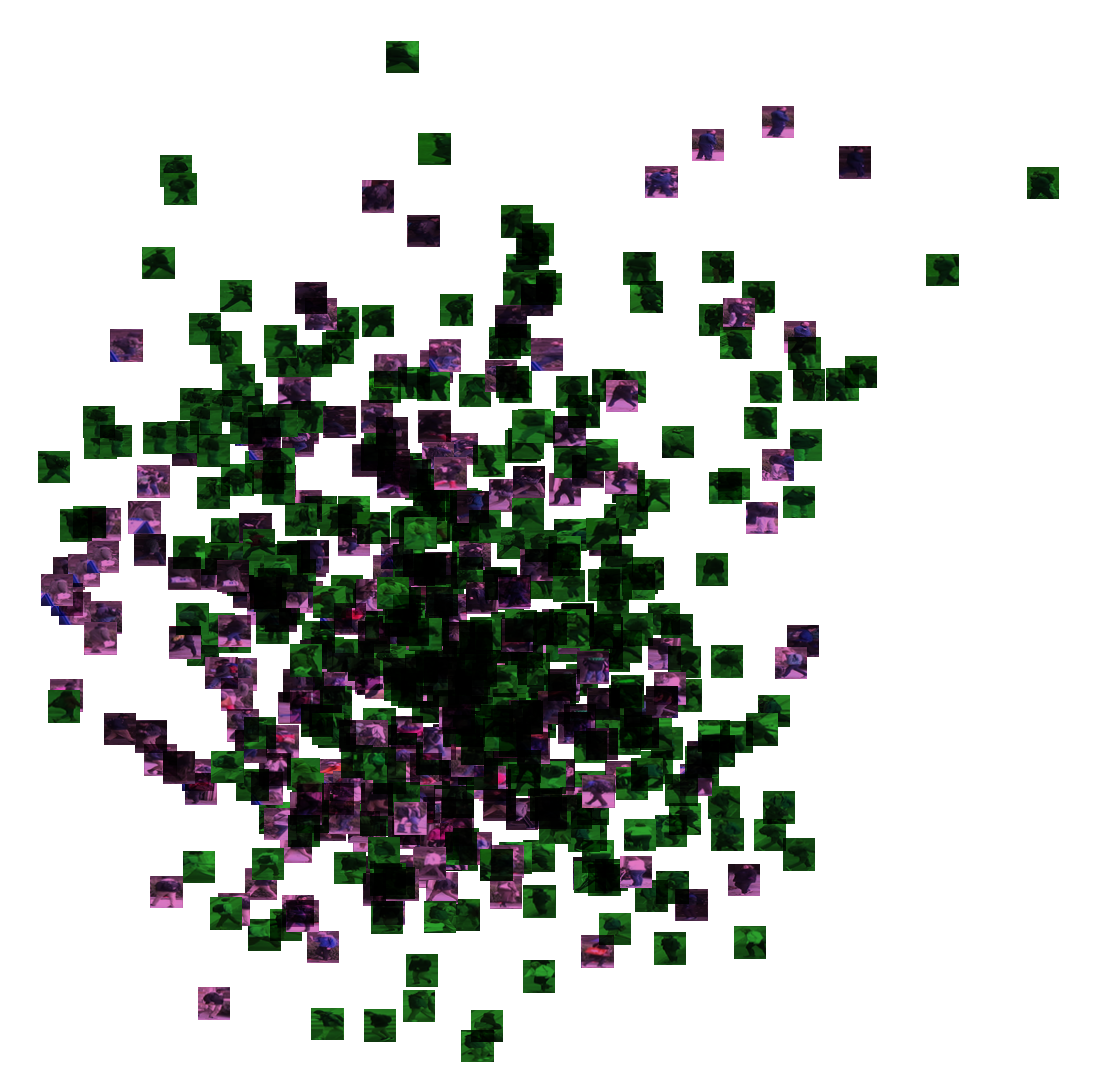}}
\subfigure[CANU-MMT, {\color{purple} 2} and {\color{green} 4}]{\includegraphics[width=0.3\textwidth]{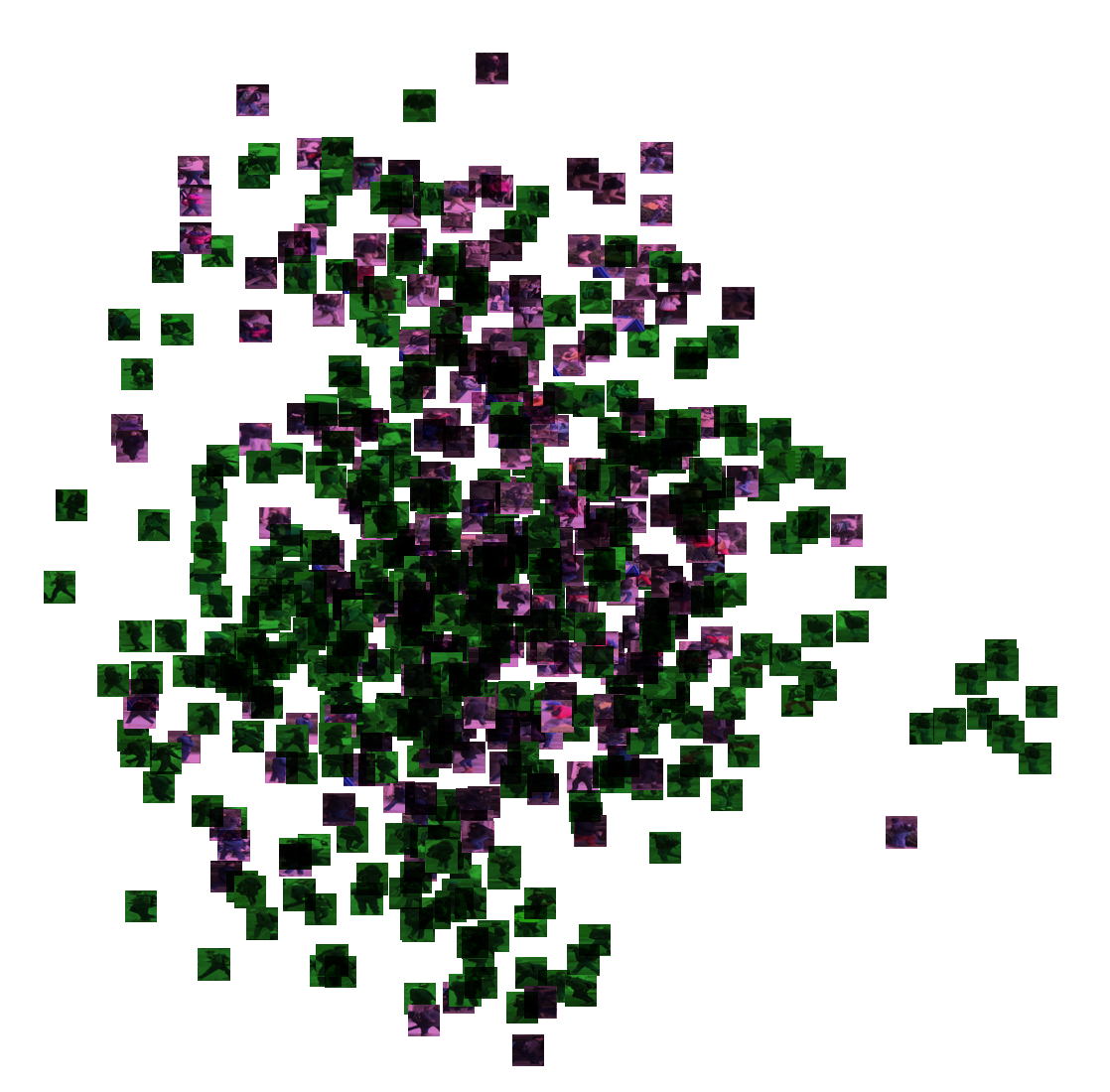}}

\subfigure[MMT, camera \textcolor{blue}{0} and \textcolor{orange}{6}]{\includegraphics[width=0.3\textwidth]{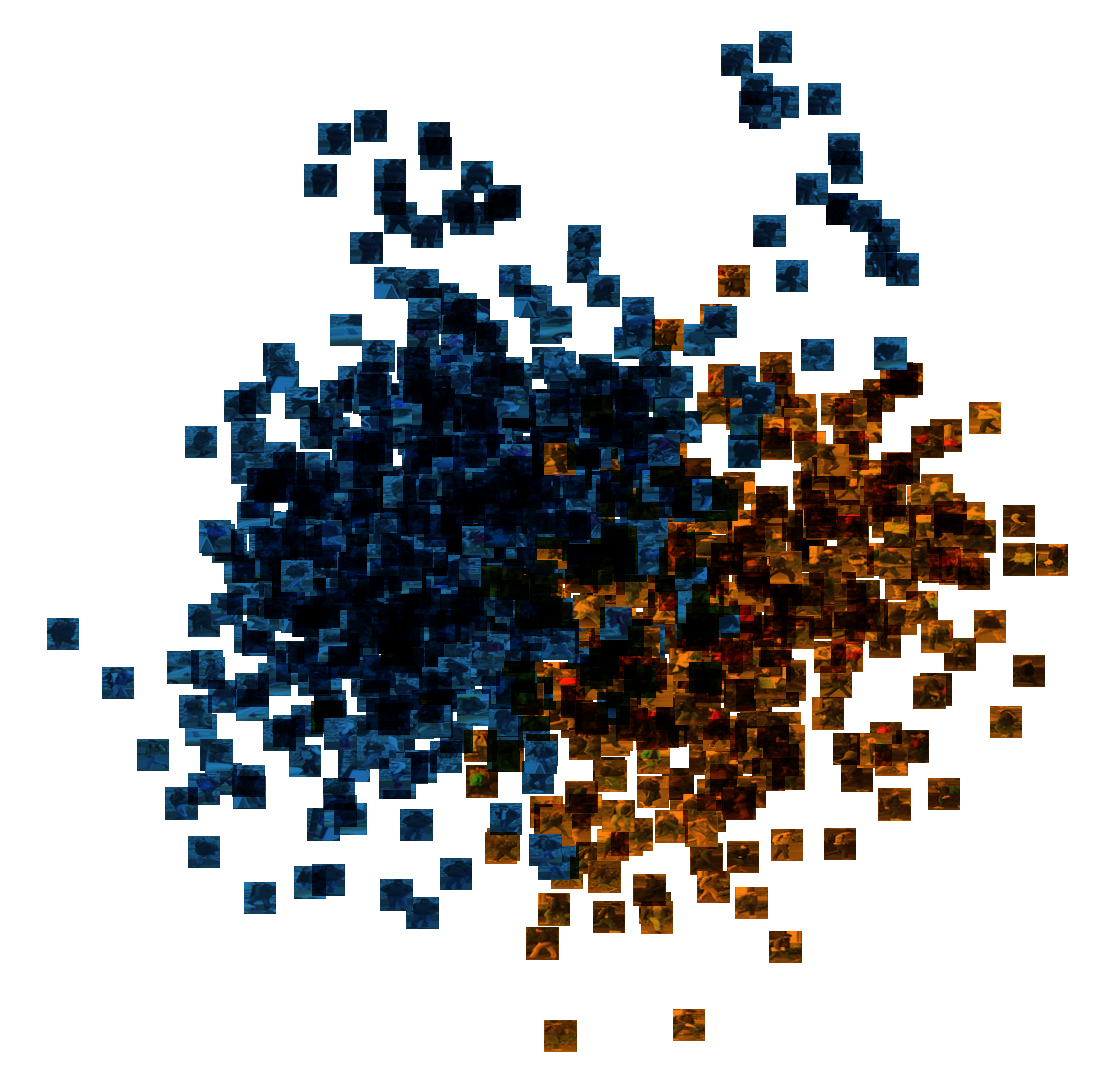}}
\subfigure[MMT+Adv, camera \textcolor{blue}{0} and \textcolor{orange}{6}]{\includegraphics[width=0.3\textwidth]{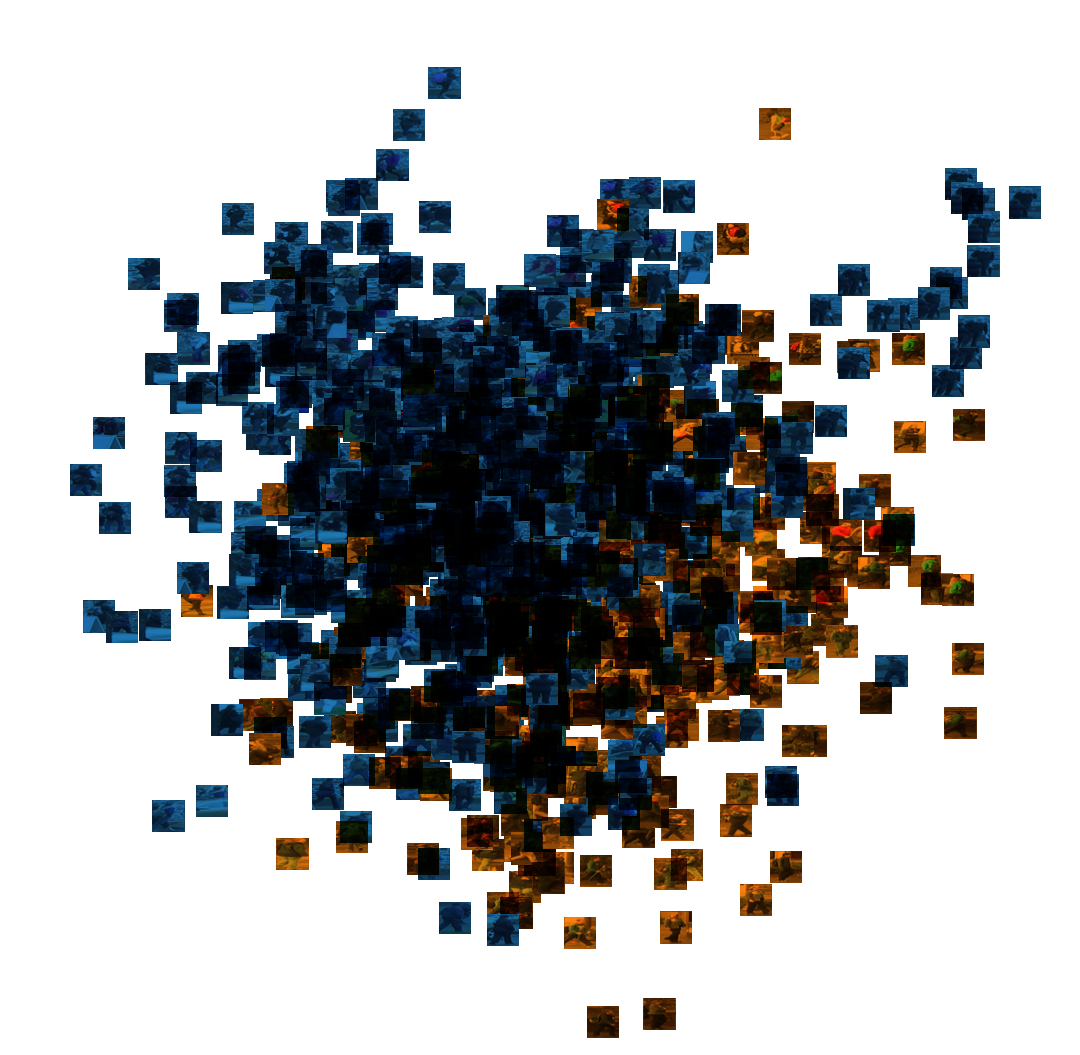}}
\subfigure[CANU-MMT, camera \textcolor{blue}{0} and \textcolor{orange}{6}]{\includegraphics[width=0.3\textwidth]{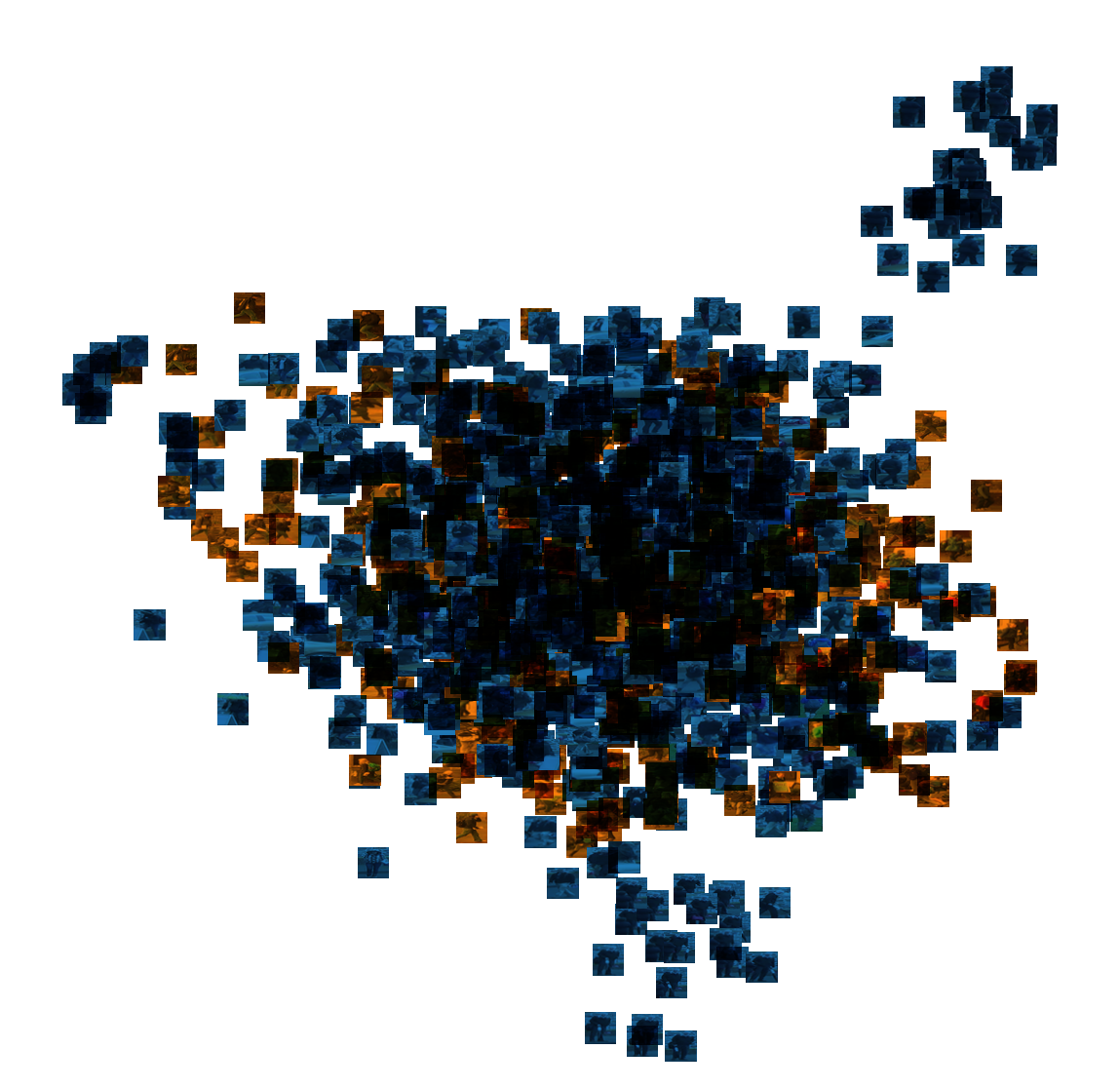}}
\vspace*{-3mm}
\caption{PCA visualization of the embedding for Mkt\myto\ Duke setting. Best viewed in color. \label{fig:visu}}
\end{figure*}

We use the feature embedding to compute a PCA projection of the embedding space for different pairs of cameras, and use the first 2 dimensions for visualization, in Figure \ref{fig:visu}. We use PCA projection to preserve the global structure, which is not guaranteed in other dimensionality reduction methods like t-SNE. We report the Mkt\myto\ Duke setting for MMT, MMT+Adv, and \method-MMT, using the train set of the target dataset. We observe that for all pairs of cameras, the embedding vectors distributions overlap more for both adversarial strategies, compared to the original implementation of MMT. It shows that our adversarial strategies achieve distributions matching across cameras more reliably than MMT alone. 

\end{document}